\documentclass[10pt,twocolumn,twoside]{IEEEtran}
\usepackage{amsmath}
\usepackage{graphicx}
\usepackage{todonotes, comment}
\usepackage{amssymb}
\usepackage{multirow}
\usepackage{xcolor}
\usepackage{algorithm}
\usepackage{cite}
\usepackage{amsthm}
\usepackage{amssymb}
\usepackage{mathtools}
\newtheorem{theorem}{Theorem}

\newtheorem{lemma}{Lemma}

\newtheorem*{remark}{Remark}
\usepackage[caption=false,font=footnotesize]{subfig}
\usepackage[noend]{algpseudocode}
\usepackage{hyperref}

\def\cast{{
   \mathord{
      \hbox to 0em{
         \ooalign{
	   \smash{\hbox{$\ast$}}\crcr
	   \smash{\hskip-1pt\Large\hbox{$\circ$}} }
	 \hidewidth}
      \phantom{\bigcirc}
} }}

\newcommand{\rT}{^{ \raisebox{1.2pt}{$\rm \scriptstyle T$}}}

\newcommand{\bds}{\begin {itemize}}
\newcommand{\eds}{\end {itemize}}
\newcommand{\bdf}{\begin{definition}}
\newcommand{\blm}{\begin{lemma}}
\newcommand{\edf}{\end{definition}}
\newcommand{\elm}{\end{lemma}}
\newcommand{\bthm}{\begin{theorem}}
\newcommand{\ethm}{\end{theorem}}
\newcommand{\bprp}{\begin{prop}}
\newcommand{\eprp}{\end{prop}}
\newcommand{\bcl}{\begin{claim}}
\newcommand{\ecl}{\end{claim}}
\newcommand{\bcr}{\begin{coro}}
\newcommand{\ecr}{\end{coro}}
\newcommand{\bquest}{\begin{question}}
\newcommand{\equest}{\end{question}}

\newcommand{\larrow}{{\larrow}}

\newcommand{\argmin}{\ensuremath{\mathrm{arg}\min}}
\newcommand{\argmax}{\ensuremath{\mathrm{arg}\max}}

\newcommand{\cC}{{\ensuremath{\mathcal{C}}}}

\newcommand{\cE}{{\ensuremath{\mathcal{E}}}}

\newcommand{\cG}{{\ensuremath{\mathcal{G}}}}

\newcommand{\cL}{{\ensuremath{\mathcal{L}}}}

\newcommand{\cN}{{\ensuremath{\mathcal{N}}}}

\newcommand{\cP}{{\ensuremath{\mathcal{P}}}}

\newcommand{\cS}{{\ensuremath{\mathcal{S}}}}

\newcommand{\cV}{{\ensuremath{\mathcal{V}}}}

\newcommand{\va}{{\ensuremath{{\mathbf{a}}}}}

\newcommand{\vb}{{\ensuremath{{\mathbf{b}}}}}
\newcommand{\vc}{{\ensuremath{{\mathbf{c}}}}}

\newcommand{\vq}{{\ensuremath{{\mathbf{q}}}}}

\newcommand{\vx}{{\ensuremath{{\mathbf{x}}}}}

\newcommand{\mA}{{\ensuremath{\mathbf{A}}}}

\newcommand{\mC}{{\ensuremath{\mathbf{C}}}}

\newcommand{\mF}{{\ensuremath{\mathbf{F}}}}

\newcommand{\mI}{{\ensuremath{\mathbf{I}}}}

\newcommand{\mP}{{\ensuremath{\mathbf{P}}}}

\newcommand{\mS}{{\ensuremath{\mathbf{S}}}}

\newcommand{\mX}{{\ensuremath{\mathbf{X}}}}

\usepackage{latexsym}

\def\IC{\mathbb C}
\def\IN{\mathbb N}
\def\IZ{\mathbb Z}
\def\IR{\mathbb R}

\def\shat{^{\mathchoice{}{}%
 {\,\,\smash{\hbox{\lower4pt\hbox{$\widehat{\null}$}}}}%
 {\,\smash{\hbox{\lower3pt\hbox{$\hat{\null}$}}}}}}

\def\bSigma{{
      \ooalign{
      \smash{\hskip.4pt\raise.4pt\hbox{$\Sigma$}}\vphantom{}\crcr
      \smash{\hskip.7pt\raise.6pt\hbox{$\Sigma$}}\vphantom{}\crcr
      \smash{\hbox{$\Sigma$}}\vphantom{$\Sigma$}}
      \vphantom{\hbox{$\Sigma$}}
      }}
\def\bTheta{{
      \ooalign{
      \smash{\hskip.5pt\raise.5pt\hbox{$\Theta$}}\vphantom{}\crcr
      \smash{\hskip.0pt\raise.1pt\hbox{$\Theta$}}\vphantom{}\crcr
      \smash{\hbox{$\Theta$}}\vphantom{$\Theta$}}
      \vphantom{\hbox{$\Theta$}}
      }}
\def\bDelta{{
      \ooalign{
      \smash{\hskip.4pt\raise.4pt\hbox{$\Delta$}}\vphantom{}\crcr
      \smash{\hskip.7pt\raise.6pt\hbox{$\Delta$}}\vphantom{}\crcr
      \smash{\hbox{$\Delta$}}\vphantom{$\Delta$}}
      \vphantom{\hbox{$\Delta$}}
      }}
\def\bLambda{{
      \ooalign{
      \smash{\hskip.5pt\raise.5pt\hbox{$\Lambda$}}\vphantom{}\crcr
      \smash{\hskip.0pt\raise.1pt\hbox{$\Lambda$}}\vphantom{}\crcr
      \smash{\hbox{$\Lambda$}}\vphantom{$\Lambda$}}
      \vphantom{\hbox{$\Lambda$}}
      }}

\makeatletter

\def\bordermatrix#1{\begingroup \m@th
  \@tempdima 8.75\p@
  \setbox\z@\vbox{%
    \def\cr{\crcr\noalign{\kern2\p@\global\let\cr\endline}}%
    \ialign{$##$\hfil\kern2\p@\kern\@tempdima&\thinspace\hfil$##$\hfil
      &&\quad\hfil$##$\hfil\crcr
      \omit\strut\hfil\crcr\noalign{\kern-\baselineskip}%
      #1\crcr\omit\strut\cr}}%
  \setbox\tw@\vbox{\unvcopy\z@\global\setbox\@ne\lastbox}%
  \setbox\tw@\hbox{\unhbox\@ne\unskip\global\setbox\@ne\lastbox}%
  \setbox\tw@\hbox{$\kern\wd\@ne\kern-\@tempdima\left[\kern-\wd\@ne
    \global\setbox\@ne\vbox{\box\@ne\kern2\p@}%
    \vcenter{\kern-\ht\@ne\unvbox\z@\kern-\baselineskip}\,\right]$}%
  \null\;\vbox{\kern\ht\@ne\box\tw@}\endgroup}
\makeatother

\makeatletter
\def\argmin{\mathop{\operator@font arg\,min}}
\def\argmax{\mathop{\operator@font arg\,max}}
\makeatother

\newcommand{\tr}{\mbox{\rm tr}}

\newcommand{\bea}{\begin{array}}
\newcommand{\ena}{\end{array}}
\newcommand{\beq}{\begin{equation}}
\newcommand{\enq}{\end{equation}}

\newcommand{\beqa}{\begin{eqnarray}}
\newcommand{\enqa}{\end{eqnarray}}

\newcommand{\beqan}{\begin{eqnarray*}}
\newcommand{\enqan}{\end{eqnarray*}}

\newcommand{\AL}{\begin{enumerate}}
\newcommand{\ALE}{\end{enumerate}}

\def\addots{\mathinner{
    \mkern1mu\raise0pt\vbox{\kern7pt\hbox{.}}
    \mkern2mu\raise4pt\hbox{.}
    \mkern2mu\raise7pt\hbox{.}
    \mkern1mu}}

\def\sddots{\mathinner{
    \mkern.8mu\raise7pt\hbox{.}
    \mkern.8mu\raise4pt\hbox{.}
    \mkern.8mu\raise0pt\vbox{\kern7pt\hbox{.}}
    \mkern1mu}}

\def\saddots{\mathinner{
    \mkern.2mu\raise0pt\vbox{\kern7pt\hbox{.}}
    \mkern.2mu\raise4pt\hbox{.}
    \mkern.2mu\raise7pt\hbox{.}
    \mkern1mu}}

\def\sqplus{\mathbin{
	{\ooalign{\hfil\raise.3ex\hbox{\scriptsize
	+}\hfil\crcr\mathhexbox274\crcr\mathhexbox275}}
	}} 
\def\sqminus{\mathbin{
	{\ooalign{\hfil\raise.3ex\hbox{\scriptsize
	--}\hfil\crcr\mathhexbox274\crcr\mathhexbox275}}
	}}

\def\IC{{
   \mathord{
      \hbox to 0em{
	 \hskip-4pt
         \ooalign{
	   \smash{\hskip1.9pt\raise2.6pt\hbox{$\scriptscriptstyle |$}}\crcr
	   \smash{\hbox{\rm\sf C}} }
	 \hidewidth}
      \phantom{\hbox{\rm\sf C}}
} }}
\def\IN{
    {\ooalign{
   \smash{\hskip2.2pt\raise1.5pt\hbox{$\scriptscriptstyle |$}}\vphantom{}\crcr
   \hbox{\sf N}
	}}
	} 
\def\IZ{
    {\ooalign{
   \smash{\hskip1.9pt\raise0pt\hbox{$\sf Z$}}\vphantom{}\crcr
   \hbox{\sf Z}
	}}
	} 
\def\IR{
    {\ooalign{
   \smash{\hskip2.2pt\raise1.5pt\hbox{$\scriptscriptstyle |$}}\vphantom{}\crcr
   \smash{\hskip2.2pt\raise3.3pt\hbox{$\scriptscriptstyle |$}}\vphantom{}\crcr
   \hbox{\sf R}
	}}
	} 

\DeclareMathAlphabet{\mathcmb}{OT1}{cmr}{b}{n}

\def\bSigma{\ensuremath{\mathcmb{\Sigma}}}
\def\bLambda{\ensuremath{\mathcmb{\Lambda}}}

\def\bTheta{\ensuremath{\mathcmb{\Theta}}}

\newcommand{\SI}{\begin{indlist}}
\newcommand{\EI}{\end{indlist}}

\newcommand{\DL}{\begin{dashlist}}
\newcommand{\DLE}{\end{dashlist}}

\makeatletter
\def\setboxz@h{\setbox\z@\hbox}
\def\wdz@{\wd\z@}
\def\boxz@{\box\z@}
\def\underset#1#2{\binrel@{#2}%
  \binrel@@{\mathop{\kern\z@#2}\limits_{#1}}}
\def\binrel@#1{\begingroup
  \setboxz@h{\thinmuskip0mu
    \medmuskip\m@ne mu\thickmuskip\@ne mu
    \setbox\tw@\hbox{$#1\m@th$}\kern-\wd\tw@
    ${}#1{}\m@th$}%
  \edef\@tempa{\endgroup\let\noexpand\binrel@@
    \ifdim\wdz@<\z@ \mathbin
    \else\ifdim\wdz@>\z@ \mathrel
    \else \relax\fi\fi}%
  \@tempa
}
\let\binrel@@\relax%
\makeatother

\begin{document}

\title{Generative Models and Learning Algorithms for Core-Periphery Structured Graphs}
\author{Sravanthi~Gurugubelli,~\IEEEmembership{Student Member,~IEEE},~and~Sundeep~Prabhakar Chepuri,~\IEEEmembership{Member,~IEEE}
\thanks{S. Gurugubelli and S.P. Chepuri are with the Department of ECE, Indian Institute of Science, Bangalore, India. Email: \{sravanthig;spchepuri\}@iisc.ac.in}
}

\maketitle

\begin{abstract}
We consider core-periphery structured graphs, which are graphs with a group of densely and sparsely connected nodes, respectively, referred to as core and periphery nodes. The so-called core score of a node is related to the likelihood of it being a core node. In this paper, we focus on learning the core scores of a graph from its node attributes and connectivity structure. To this end, we propose two classes of probabilistic graphical models: affine and nonlinear. First, we describe affine generative models to model the dependence of node attributes on its core scores, which determine the graph structure. Next, we discuss nonlinear generative models in which the partial correlations of node attributes influence the graph structure through latent core scores. We develop algorithms for inferring the model parameters and core scores of a graph when both the graph structure and node attributes are available. When only the node attributes of graphs are available, we jointly learn a core-periphery structured graph and its core scores. We provide results from numerical experiments on several synthetic and real-world datasets to demonstrate the efficacy of the developed models and algorithms. 
\end{abstract}

\IEEEpeerreviewmaketitle

\section{Introduction \label{sec:Introduction}}

\begin{figure}
	\centering
	\includegraphics[width=\columnwidth]{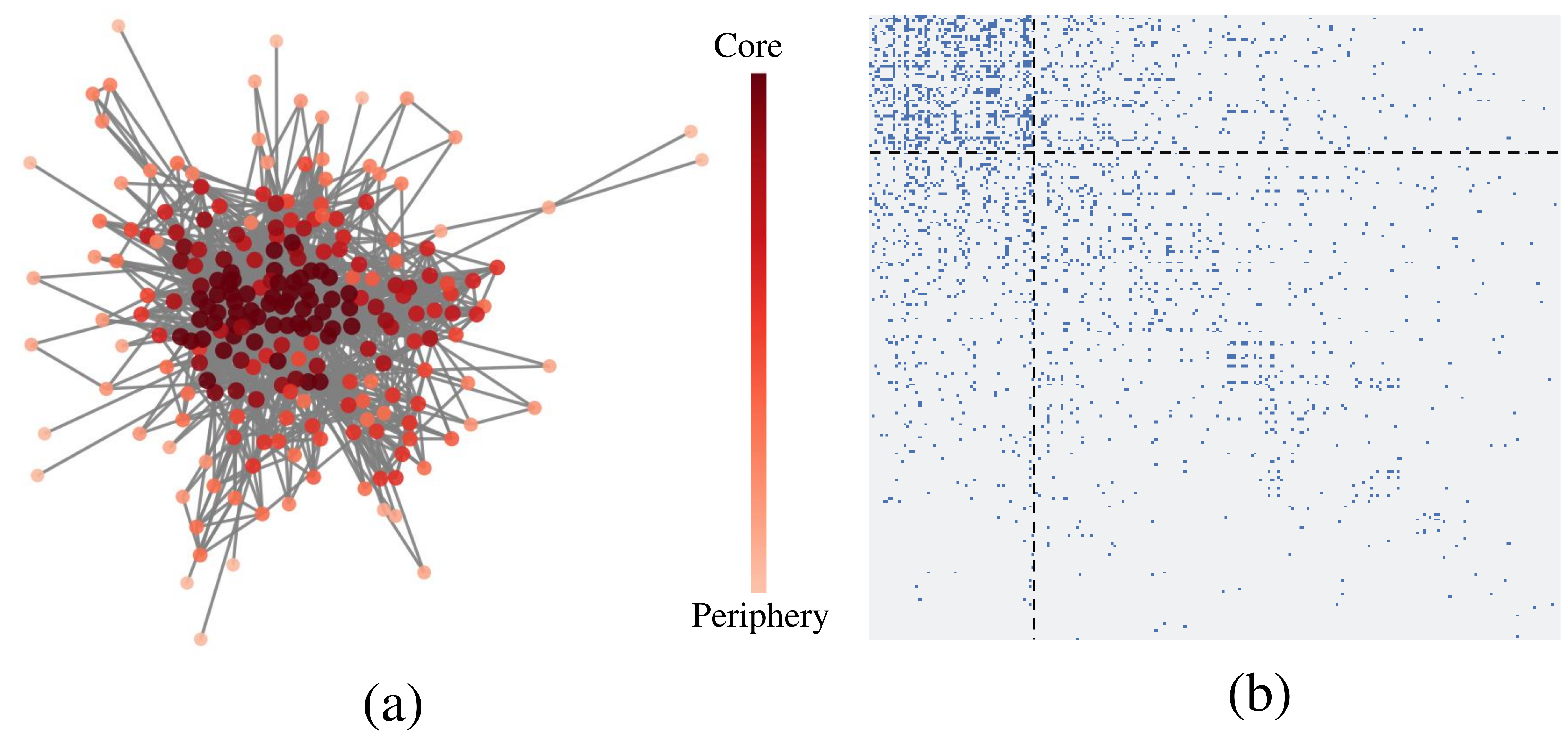}
	\caption{(a) A network with a core-periphery structure. The dark colored nodes are the core nodes while the lighter ones form the periphery nodes. (b)  Its adjacency matrix with rows and columns ordered according to the decreasing nodal core score.
	}
	\label{fig:cpexample}
	\vspace{-4mm}
\end{figure}

\IEEEPARstart{C}{ore}-periphery structured graphs have densely connected groups of nodes, called \emph{core nodes}, and sparsely connected groups of nodes, called \emph{periphery nodes}. While the core nodes are cohesively connected to the other core nodes and are also reasonably well connected to the peripheral nodes, the peripheral nodes are not well connected to each other or to any core node in the graph. Core-periphery structured graphs are ubiquitous and are extensively used to analyze real-world networks such as social networks~\cite{barbera2015the}, trade and transport networks~\cite{Verma2016Emergence}, and brain networks~\cite{bassett2013task}, to name a few. Identifying the core nodes in a network allows us to analyze crucial processes in it. For instance, in brain networks, atypical core-periphery brain dynamics are observed in subjects with autism spectrum disorder~\cite{Harlalka2019Atypical}. In social networks~\cite{barbera2015the} (contact networks~\cite{Kitsak2010Identification}), the most influential spreaders of information (respectively, a disease) are usually the core nodes. An example of a core-periphery structured social network of a subset of $244$ Twitter users~\cite{leskovec2012Learning} is shown in Fig.~\ref{fig:cpexample}. 

Given an adjacency matrix of a graph, its rows and columns can be permuted to reveal the underlying core-periphery structure.  However, generating all the possible permutations of its rows and columns to arrive at an adjacency matrix that reveals the core-periphery structure is an NP-hard problem.  To this end, many heuristics that reveal the core-periphery structure from an adjacency matrix are available~\cite{Borgatti2000Models, Rombach2014Core, Boyd2010Computing, Alvarez2005K, Della2013Profiling,Jia2019Random}.  Another commonly used approach to identify the core nodes in a graph is by learning the so-called \emph{core score} of a node, where the core score of a node is related to the likelihood of that node belonging to the core part of the graph.  In some networks, such as trade, transport, or brain networks, the core score of a node also depends on its spatial distance from the other nodes~\cite{Jia2019Random}.  Specifically, in such networks, a node that is spatially far away from another node is very unlikely to be connected to it.  For the graph in Fig.~\ref{fig:cpexample}(a), a permuted adjacency matrix (permuted according to the rank-ordered core scores) is shown in Fig.~\ref{fig:cpexample}(b), wherein the dense entries in the top left block correspond to the core-core connections in the graph and the sparse entries in the bottom right block correspond to the periphery-periphery connections.

In many real-world networks, we only have access to node attributes, and the underlying graph might not always be available. For example, in a brain network, we usually have access to functional magnetic resonance imaging (fMRI) data, but the structural connectivity information is often unavailable. Node attributes carry vital information that often complements the information in the graph structure and can improve the quality of the core score estimates. However, existing works~\cite{Borgatti2000Models, Rombach2014Core, Boyd2010Computing, Alvarez2005K, Della2013Profiling,Jia2019Random} rely only on the knowledge of the underlying graph and do not use node attributes for inferring the core scores. Therefore, this work proposes probabilistic models and algorithms to infer the core scores  in the following two cases, namely, $(i)$ when both the underlying graph structure and node attributes are available and $(ii)$ when only node attributes are available.

\subsection{Relevant Prior Works \label{subsec:PriorWorks}}
Existing works infer the core scores from a graph, i.e., from an adjacency matrix. In~\cite{Borgatti2000Models}, correlation measures to quantify how well a network approximates the ideal model of a core-periphery structured graph are defined, where the ideal model consists of a core-core block that is fully connected, a periphery-periphery block with no connections, and a core-periphery part that is either fully connected or not connected at all. These measures are then used to develop algorithms to estimate the core scores. In~\cite{Rombach2014Core}, the core score vector is learnt by maximizing the correlation measure from~\cite{Borgatti2000Models} with a constraint that the core score vector is a shuffled version of an $N$-dimensional vector whose entries are fixed according to a number of desired factors that effect the spread of the core scores, such as the number of nodes in the core region and the change in the core score from the core to peripheral nodes. In~\cite{Boyd2010Computing}, the edge weight of a node pair is modeled as the product of its cores scores, and the core score vector is obtained by minimizing the sum of the squared differences between the known edge weights and the product of the core scores of the related node pair, summed over all node pairs. In~\cite{Alvarez2005K}, a recursive algorithm, called the k-core decomposition, is proposed to iteratively partition a graph into nodes having sparse and dense connections. In~\cite{Della2013Profiling}, an iterative algorithm derived from a standard random walk model is proposed, where the so-called persistence probability that denotes the probability that a random walker starting from a certain node in a sub-network of a network remains in that sub-network in the next step is defined. Then starting from a sub-network containing a single node with the weakest connectivity, nodes are iteratively added to the sub-network such that the increase in persistence probability is minimal. The above mentioned existing works, which we refer as, \texttt{Rombach}~\cite{Rombach2014Core}, \texttt{MINRES}~\cite{Boyd2010Computing}, \texttt{k-cores}~\cite{Alvarez2005K}, and \texttt{Random-Walk}~\cite{Della2013Profiling}  serve as the baseline for the proposed methods.

When only node attributes are available and the underlying graph structure is unknown, we first need to learn the underlying graph. We may then infer the core scores using existing methods. One of the standard approaches to infer graphs from node attributes is graphical lasso, which solves an $\ell_1$-regularized Gaussian maximum log-likelihood problem~\cite{Friedman2008Sparse} to learn a sparse connectivity pattern underlying a Gaussian graphical model. However, graphical lasso imposes a uniform $\ell_1$ penalty on each edge and does not account for the core-periphery structure in networks.

To summarize, when we have access to only node attributes and the underlying graph is unavailable, the existing methods cannot be used to estimate core scores. Furthermore, the existing works do not incorporate node attribute information while inferring core scores. In this work, we propose generative models and learning algorithms to address these limitations. 

\subsection{Main Results and Contributions \label{subsec:MainResultsandContributions}}
The main results and contributions of the paper are summarized as follows.

\begin{itemize}
	\item \textbf{Generative Models}: We propose probabilistic generative models that relate core-periphery structured graphs to their core scores and attributes of nodes to their respective core scores through an affine model. We refer to these models as \texttt{Graph-Attributes-Affine (GA-Affine)}. In particular, we propose two models, namely, \texttt{GA-Affine-Bool} and \texttt{GA-Affine-Real}, to account for binary and real-valued node attributes, respectively. Next, we propose nonlinear generative models that, though not as simple as the affine models, incorporate information about spatial distances and capture any dependencies between node attributes. Within the nonlinear models, we propose two models, namely, \texttt{Graph-Attributes-Nonlinear (GA-Nonlinear)} and \texttt{Attributes-Only (AO)} to address two different cases, namely, both node attributes and graph structure being available and only the node attributes being available, respectively. Similar to the \texttt{GA-Affine} model, the \texttt{GA-Nonlinear} model assumes the graph is known. However, contrary to the \texttt{GA-Affine} model, it establishes a nonlinear relationship between the core scores and the node attributes. On the other hand, the nonlinear \texttt{AO} model assumes that the underlying graph is unknown and relates node attributes to core scores through a latent graph structure. In~\cite{Gurugubelli2022Learning}, which is a conference precursor of this work, we proposed the \texttt{AO} model, which we extend to generative models that relate core scores to graphs and node attributes in the paper. 
	
	\item \textbf{Algorithms}: We infer core scores using both the graph structure and the complementary information contained in the node attributes whenever both are available by fitting one of the proposed models of \texttt{Graph-Attributes} to the observed data. The problems formulated using models \texttt{GA-Affine-Bool} and \texttt{GA-Affine-Real} are non-convex in the core scores and model parameters, which we solve by alternating minimization. With \texttt{GA-Nonlinear}, the inference problem is a convex problem in the core scores for which we propose a projected gradient ascent based solver. Next, when only node attributes are available, we fit data to the \texttt{AO} model to simultaneously learn the unknown latent graph structure and the core scores. The inference problem takes the form of the graphical lasso problem. However, in contrast to the standard graphical lasso, the $\ell_1$ regularization is not uniformly applied on all the edges but is weighed according to the latent core scores. We provide an alternating minimization based solver to jointly learn the underlying graph and the core scores.
	
\end{itemize} 

We test the proposed algorithms on several synthetic and real-world datasets. For synthetic datasets, we observe that the core scores estimated by the proposed algorithms are the closest to the groundtruth core scores, in terms of cosine similarity, compared to those estimated by the existing methods that consider only the graph structure. We also perform various numerical experiments to show the ability of the proposed algorithms to correctly identify the core parts of various real-world datasets such as citation, organizational, social, and transportation networks. Further, when the graph structure is unavailable, we show that our method learns the core scores along with the core-periphery structured graphs outperforming graphical lasso. We also use as inputs the core scores and graphs learnt using the algorithm obtained from the proposed \texttt{AO} model to classify healthy individuals and individuals with attention deficit hyperactivity disorder (ADHD) using graph neural networks when only the fMRI data of the subjects is available and also illustrate the major differences in the core and periphery regions of the two groups. Software and data to reproduce the results in the paper is available at~\url{https://github.com/SravanthiGurugubelli/CPGraphs}.

\subsection{Notation and Organization \label{subsec:NotationandOrganization}}
Throughout the paper, boldface lowercase (uppercase) letters denote column vectors (respectively, matrices). Operators ${\rm tr}(.)$, $(.)^{-1}$, and $(.)\rT$ stand for trace, inverse, and transpose operations, respectively. $\mI_N$ is the identity matrix of size $N\times N$.  $\mathbf{1}_{mn}$ and $\mathbf{0}_{mn}$ denote $m \times n$ dimensional matrices with all ones and all zeros, respectively. The projection operator to project a vector $\vc=[c_1,c_2,\cdots,c_N]\rT\in \mathbb{R}^N$ onto a constraint set formed by the intersection of a hyperplane $\sum_{i=1}^N c_i = M$ with $M$ being a positive constant and a rectangle $\vc \in [0,1]^N$ is given by:
\begin{align}
	P_{\cC}\left\{\vc^{(t)}\right\} = P_{[0,1]}\left\{\vc^{(t)}-\lambda^*\mathbf{1}\right\},
	\label{eq:prox}
\end{align}
where $\lambda^*\in \mathbb{R}$ is the solution to 
\begin{align}
	\phi(\lambda) = \mathbf{1}\rT P_{[0,1]}\left\{\vc-\lambda\mathbf{1}\right\}-M=0. \label{eq:lambda}
\end{align}
Since the function $\phi(\lambda)$ is a non-increasing function of $\lambda$, the root of~\eqref{eq:lambda}, denoted by $\lambda^*$, can be easily found using simple techniques like the bisection method. Here, $P_{[0,1]}\left\{\va\right\}$ denotes the projection operator for projecting the vector $\va=[a_1,a_2,\cdots,a_n]\rT$ onto the rectangle $\va \in [0,1]^N$ and is given by the component-wise operator
$$
{\left[P_{[0,1]}\left\{\va\right\}\right]}_{k}= \begin{cases}0, & a_{k} \leq 0, \\ a_{k}, & 0 \leq a_{k} \leq 1, \\ 1, & a_{k} \geq 1.\end{cases}
$$

The rest of the paper is organized as follows. In Section~\ref{sec:graph on cs}, we model the dependence of graph structures on core scores. In Section~\ref{sec:affine models}, we model the dependence of node attributes on their respective core scores through affine functions and develop algorithms to infer core scores from both the graph and its node attributes. In Section~\ref{sec:nonlinear models}, we propose nonlinear models for node attributes and develop algorithms to infer core scores in the following two cases: when both node attributes and graph are available and when only node attributes are available. In Section~\ref{sec:NumericalExp}, we present results from several numerical experiments to evaluate the proposed algorithms in terms of their ability to correctly identify the core scores of several synthetically generated and real-world networks. Finally, we conclude the paper in Section~\ref{sec:conclusions}.

\section{Modeling Core Scores \label{sec:graph on cs}}
Consider a weighted and undirected graph $\cG= \{\cV,\cE\}$, where $\cV = \{v_1, \cdots v_N\}$ is the vertex set with $N$ vertices and $\cE$ is the edge set. The connectivity structure of $\cG$ is captured in $\bTheta \in \mathbb{R}^{N\times N}$, i.e., the $(i,j)$th entry of $\bTheta$ denoted by $\Theta_{ij}$ is nonzero if nodes $i$ and $j$ are connected and is zero otherwise. Let us denote the node attribute data as $\mX = [ \vx_1, \vx_2, \cdots, \vx_d ] \in \mathbb{R}^{N\times d}$ with the 
$i$th row of $\mX$ containing $d$-dimensional features of the entity associated to the $i$th node of $\cG$. Let $\vc=[c_1,c_2,\cdots,c_N]\rT \in \mathbb{R}_{+}^{N}$ denote the core score vector of a graph $\cG$ with $N$ nodes, where the $i$th entry $c_i \in [0,1]$ denotes the coreness value of node $i$. The core score $c_i$ denotes the likelihood of node $i$ belonging to the core part of the graph with $c_i=1$ denoting the certainty of node $i$ belonging to the core part of the graph. We denote the spatial distance between nodes $i$ and $j$ by $d_{ij}$. 

In this section, we develop a model to relate the graph $\bTheta$ to the core scores of its nodes. Specifically, we model $\bTheta$ such that it induces a sparsity pattern in graphs determined by the core scores of its nodes and the spatial distances between the nodes. The proposed coreness and position-aware probabilistic model on $\bTheta$ is based on the following intuitions. Firstly, the connections between nodes with large core scores are dense, between nodes with large and small core scores are relatively sparser, and between nodes with small core scores are very sparse. Secondly, spatially distant nodes have sparse connections between them. 

Let us define a parameter, $e$, which determines the dependence of $\Theta_{ij}$ on spatial distances relative to the dependence on core scores in the model. With a small constant $\epsilon$, we model $\Theta_{ij}$ such that when the value $c_i+c_j-e{\rm log}(d_{ij}+\epsilon)$ for any two nodes $i$ and $j$ is large, $\Theta_{ij}$ is large. When spatial distances between nodes are not relevant or are not available, $e$ is simply set to zero. To satisfy the above mentioned requirements, we model the entries of $\bTheta$ as Laplace random variables with \[w_{ij}=1-c_i-c_j+e{\rm log}(d_{ij}+\epsilon)\] being the inverse diversity parameter, which depends on the latent variables $c_i$, $c_j$, and the spatial distances $d_{ij}$ for $i,j=1,\cdots,N$. In other words, the probability density function (PDF) of $\bTheta$ parameterized by the latent core score vector $\vc$ is
\begin{align}
	p(\bTheta;\vc) &= \prod_{i,j=1}^{N}p(\Theta_{ij};c_i,c_j)  \nonumber\\
	&\propto \prod_{i,j=1}^{N} {\rm exp}\left(-\lambda w_{ij}|\Theta_{ij}|\right),
	\label{eq:theta;c}
\end{align} 
where $\lambda>0$ controls the dependence of $\bTheta$ on the core scores and the spatial distances. For $p(\bTheta;\vc)$ to be a valid probability distribution, $w_{ij}$ should be positive.

Throughout the paper, we employ the probabilistic model for $\bTheta$ from~\eqref{eq:theta;c}, where we model $\Theta_{ij}$ as a random variable drawn from a Laplace distribution with inverse diversity parameter $w_{ij}$. Now that we have a model that relates the graph structure to the core scores, in what follows, we propose probabilistic models to relate node attributes to the core scores. We then propose algorithms to learn the core scores $\vc$ when both the graph $\bTheta$ and node attributes $\mX$ are available and when only $\mX$ is available. We first consider simple affine models to relate attributes of nodes to their respective core scores. Although affine models are simple, in some cases, the core scores of different nodes might be dependent not just on their respective node attributes but on the correlations between the node attributes or on spatial distance information. To that end, we also propose nonlinear models. 

\section{Affine Models\label{sec:affine models}}
In this section, we model the dependence of node attributes on the core scores through affine relations and develop algorithms to infer core scores using both the graph and its node attributes as inputs. We refer to this class of models as \texttt{Graph-Attributes-Affine (GA-Affine)}. We propose two generative models, namely, \texttt{GA-Affine-Bool} and \texttt{GA-Affine-Real} to generate binary-valued and real-valued node attributes from $\vc$, respectively. 
\begin{figure*}[t]
	\centering
	\includegraphics[width=2\columnwidth]{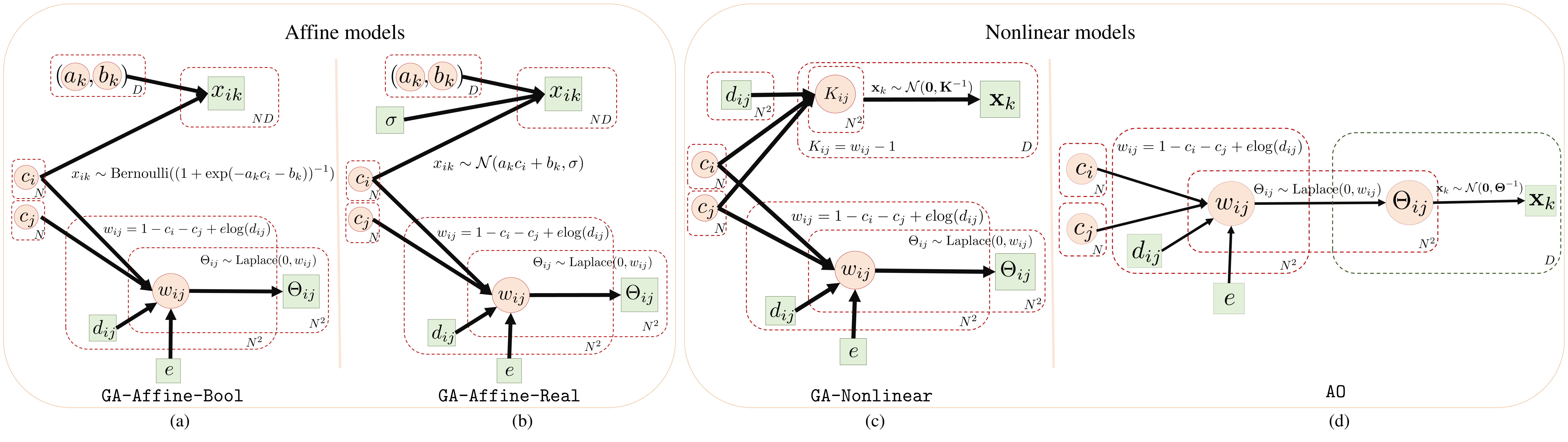}
	\caption{Plate notation for models (a) \texttt{GA-Affine-Bool}, (b) \texttt{GA-Affine-Real}, (c) \texttt{GA-Nonlinear}, and (d) \texttt{AO}. The numbers in the bottom corners of the plates indicate that the variables inside it are repeated those many times. The observed variables are shown in green squares and the latent variables in red circles. Arrows model dependencies between the variables. An arrow from a variable $a$ to another variable $b$ denotes the dependence of $b$ on $a$, and the dependence is given by the text over the arrow. In all the models, the dependence of $\bTheta$ on $\vc$ is modeled as in~\eqref{eq:theta;c}. In the \texttt{GA-Affine} models, the core scores of nodes depend affinely on their respective attributes. In model \texttt{GA-Nonlinear}, the dependence of the core scores of two nodes is modeled through the partial correlation between the attributes of the two nodes, and in the \texttt{AO} model, the dependence of the core scores of two nodes on their node attributes is modeled through the latent edges connecting them.
	} 
	\label{fig:modelOG}
\end{figure*} 
\subsection{Binary-Valued Node Attributes \label{subsubsec:BinaryAttributes}}
We first consider the case of binary node attributes. We consider a logistic model for each attribute $x_{ik}$ for $i=1,\cdots,N$, where $x_{ik}$ is $k$th binary-valued feature of node $i$. It is desired that the probability with which the model assigns the attribute $1$ to the $k$th feature of a node depends on the core score of that node through the parameters $\mF=[\va,\vb]\in\mathbb{R}^{D\times2}$, where the $k$th row of $\mF$ has entries $a_k$ and $b_k$ corresponding to the model weights of the $k$th attribute. Sharing the parameters $a_k$ and $b_k$ for $k=1,\cdots, D$ across all the nodes ensures that for feature $k$, two nodes with similar core strength are assigned the attribute $1$ with similar probability. Therefore, we model $x_{ik}$ as a Bernoulli random variable and it can be generated from the core score of that node through a logistic model as
\begin{align}
	&{\rm log}\;p(\mX;\vc,\mF) = \sum_{i=1}^{N}\sum_{k=1}^{D}{\rm log}\;p(x_{ik}; c_i,a_k,b_k) \nonumber\\
	 &= \sum_{i=1}^{N}\sum_{k=1}^{D}{x_{ik}}{\rm log}{\left(\pi_{ik}\right)} + {\left(1-x_{ik}\right)}{\rm log}{\left(1-\pi_{ik}\right)}, 
	\label{eq:X|cbinary}
\end{align}
where 
\begin{align}
	\pi_{ik}=\frac{1}{1+{\rm exp}(-a_kc_{i}-b_k)} 
	\label{eq:pi}
\end{align}
is the probability of the $k$th attribute of a node $i$ being equal to $1$. Here, $a_k$ determines the relevance of the core score of any node to the value of the $k$th node attribute. We refer to this model, which is summarized in Fig.~\ref{fig:modelOG}(a), as \texttt{GA-Affine-Bool}. Next, we develop an algorithm to learn the model parameters and the core scores from data.

\subsubsection*{Maximum Likelihood Estimator} \label{subsec:bool mle}
Given a network with binary node attributes, we infer the latent variables $\vc\in \mathbb{R}^{N}$ and $\mF\in\mathbb{R}^{D\times 2}$ of the \texttt{GA-Affine-Bool} model by maximizing the log-likelihood function  
\begin{align}
	\mathcal{L}_1(\vc,\mF) &=  {\rm log} \, p(\mX,\bTheta;\vc,\mF) \nonumber
\end{align} 
of the observed data $\left(\mX,\bTheta\right)$. Since $\mX$ and $\bTheta$ are conditionally independent given $\vc$ and $\mF$, we can express the log-likelihood as
\begin{align}
	\mathcal{L}_1(\vc,\mF)&={\rm log} \, p(\mX;\vc,\mF)+ {\rm log} \, p(\bTheta;\vc) \nonumber\\
	&= \sum\limits_{i,j=1}^{N} (c_{i}+c_{j})\lvert\Theta_{ij}\rvert  \nonumber \\ 
	&\quad\quad+ \sum_{i=1}^{N}\sum_{k=1}^{D}{x_{ik}}{\rm log}(\pi_{ik}) + (1-x_{ik}){\rm log}(1-\pi_{ik}), \nonumber
\end{align}
where ${\rm log}\,p(\bTheta;\vc)\propto \sum_{i,j=1}^N (c_i+c_j)|\Theta_{ij}|$  is as in~\eqref{eq:theta;c} with $e=0$. Ignoring the terms independent of $(\vc,\mF)$, we then have
\begin{align}
	(\mathcal{P}1): \quad \underset{ \vc\in \mathcal{C},\mF}{\text{maximize}\quad}  
	&\mathcal{L}_1(\vc,\mF) -\alpha\|\mF\|_F^2, \notag
\end{align}
where $\pi_{ik}\in [0,1]$ depends on the parameters $(c_i,a_k,b_k)$ [cf.~\eqref{eq:pi}] and the constraint set $\mathcal{C}$ is defined as 
\begin{equation}
\mathcal{C}=\left\{\vc : \sum_{i=1}^N c_i = M, c_i \in [0,1], i=1,2,\cdots, N\right\},
    \label{eq:constraintC}
\end{equation}
where $M$ is a known constant. The constraint $c_i \in [0, 1]$ fixes the scale of the core scores and the sum constraint prevents the weights $w_{ij}$, for $i,j=1,\cdots,N$ tending to zero. We regularize $\mF$ using parameter $\alpha\geq 0$ to avoid overfitting the model to noise in the input data, which is likely when the input data is scarce, i.e., when the number of nodes in the input graph is less. The problem is non-convex in the variables $\vc$ and $\mF$. We, therefore, decompose the problem into two convex sub-problems in $\vc$ and $\mF$ by fixing the other variable, respectively, and then solve them alternatingly till convergence.

\paragraph{Updating $\vc$, given $\mF$}
For a fixed $\mF$, $(\mathcal{P}1)$ simplifies to the following convex problem
\begin{align}
	\underset{ \vc\in \mathcal{C}}{\text{maximize}\quad}  
	\mathcal{L}_1(\vc,\mF)\notag
\end{align}
which we solve using projected gradient ascent. Specifically, the update for $\vc$ at iteration $t$ is given by
\begin{align}
	\vc^{(t)} = P_{\cC}\left\{\vc^{(t-1)}+\rho_c\left[2|\bTheta|\mathbf{1} + (\mX-\mP^{(t-1)})\va\right]\right\},\notag
\end{align}
where the gradient is provided in the appendix, $\rho_c$ is the step size, $(i,j)$th entry of $\mP$ is $\pi_{ij}$, and $P_{\cC}\{\cdot\}$ is the projector onto the constraint set $\cC$ [cf. Section~\ref{subsec:NotationandOrganization}]. 

\paragraph{Updating $\mF$, given $\vc$}
For a fixed $\vc$, $(\mathcal{P}1)$ reduces to 
\begin{align}
&\underset{\mF}{\text{maximize}\quad}  
\sum_{i=1}^{N}\sum_{k=1}^{D}{x_{ik}}{\rm log}(\pi_{ik})\nonumber\\
&\qquad\qquad\qquad+ (1-x_{ik}){\rm log}(1-\pi_{ik}) -\alpha \|\mF\|_F^2,\notag
\end{align}
which is an $\ell_2-$regularized logistic regression problem. We solve it using gradient ascent. Specifically, the update for $\mF$ at iteration $t$ is given by 
\begin{align}
	\mF^{(t)} = \mF^{(t-1)}+\rho_{F}\left[(\mX-\mP^{(t-1)})\rT\mC-2\alpha\mF^{(t-1)}\right],\notag
\end{align}
where $\mC=[\vc,\mathbf{1}]\in \mathbb{R}^{N\times 2}$ and the gradient is provided in the~appendix.

To summarize, the inference algorithm related to \texttt{GA-Affine-Bool} involves two alternating steps of computing $\vc$ and $\mF$ in each iteration. Updating $\vc$ constitutes several gradient ascent steps. Precomputing $|\bTheta|\mathbf{1}$ costs order $N^2$ flops and each iteration of the $\vc$-update step, excluding the precomputation step, costs approximately order $ND$ flops. Therefore, the $\vc$-update step approximately costs $k_1ND+N^2$ flops, where $k_1$ is the number of gradient ascent iterations. The $\mF$-step approximately costs $k_2ND$ flops, where $k_2$ is the number of gradient ascent iterations required to update $\mF$. Therefore, the \texttt{GA-Affine-Bool} algorithm approximately costs order $(k_1+k_2)ND+N^2$ flops per one step of the alternating minimization procedure. 

\subsection{Real-Valued Node Attributes \label{subsubsec:RealAttributes}}
For real-valued node attributes, we consider an affine relationship between node attributes and their corresponding core scores. Consider an affine model for the generation of each attribute $x_{ik}$ from $c_i$ for $i=1,\cdots,N$. We model the $k$th attribute of each node as a Gaussian random variable whose mean is an affine function of its core score. Specifically, we model $x_{ik}$ as $x_{ik}\sim\cN(a_kc_i+b_k,\sigma^2)$ and propose the following model 
\begin{align}
	p(\mX;\vc,\mF) &= \prod_{i=1}^{N}\prod_{k=1}^{D}p(x_{ik}; c_i,a_k,b_k) \nonumber\\
	&\propto \prod_{i=1}^{N}\prod_{k=1}^{D}{\rm exp}\left(\frac{-1}{\sigma^2}{\left(x_{ik}-a_kc_i-b_k\right)^2}\right),
	\label{eq:X|creal-a}
\end{align}
where the $k$th attribute of any node is approximated by an affine function of its core score with the same model parameters $a_k$ and $b_k$, implying that similar node attributes are generated for nodes with similar core scores. Here, $\sigma^2$ is the variance of the Gaussian random variables $x_{ik}$ for $i=1,\cdots,N$, $k=1,\cdots D$, and it models the spread of the node attributes from the mean. We refer to this model as \texttt{GA-Affine-Real} and is summarized as Fig.~\ref{fig:modelOG}(b).

\subsubsection*{Maximum Likelihood Estimator} \label{subsec:affine mle}
Given a network with real node attributes, we infer the latent variables $\vc\in \mathbb{R}^{N}$ and $\mF=[\va,\vb]\in\mathbb{R}^{D\times2}$ of the \texttt{GA-Affine-Real} model by maximizing the log-likelihood 
\begin{align}
	\mathcal{L}_2(\vc,\mF) = & {\rm log} \, p(\mX,\bTheta;\vc,\mF) 
	=  {\rm log} \, p(\mX;\vc,\mF)+ {\rm log} \, p(\bTheta;\vc),\nonumber \\
	&= \sum\limits_{i,j=1}^{N} (c_{i}+c_{j})\lvert\Theta_{ij}\rvert  - \|\mX-\mC\mF\rT\|_F^2, \nonumber
\end{align} 
where $p(\bTheta;\vc)\propto \sum_{i,j=1}^N |\Theta_{ij}|(c_i+c_j)$ is from~\eqref{eq:theta;c} with $e=0$ and the log-likelihood of $\mX$ from~\eqref{eq:X|creal-a} is given by \[{\rm log} \, p(\mX;\vc,\mF)=\sum_{i=1}^{N}\sum_{k=1}^{D}\frac{-1}{\sigma^2}{\left(x_{ik}-a_kc_i-b_k\right)^2}.\] We ignore the terms independent of $\vc \text{ and } \mF$ to arrive at the problem
\begin{align}
	(\mathcal{P}2): \quad &\underset{ \vc \in \mathcal{C} ,\mF}{\text{maximize}\quad}  
	\mathcal{L}_2(\vc,\mF) -\alpha \|\mF\|_F^2,\nonumber
\end{align}
where the constraint set $\cC$ is as before. We regularize $\mF$ using the parameter $\alpha\geq 0$ to avoid overfitting. The problem is non-convex in the variables $\vc$ and $\mF$. We propose to solve it using alternating maximization, wherein we update $\vc$ fixing $\mF$, and vice versa till convergence. 

\paragraph{Updating \vc, given \mF}
For fixed $\mF$, $(\mathcal{P}2)$ simplifies to 
\begin{align}
	\underset{ \vc \in \mathcal{C}}{\text{maximize}\quad}  
	&\mathcal{L}_2(\vc,\mF),\nonumber
\end{align}
which is convex in the variable $\vc$. We solve the problem by projected gradient ascent. The update for $\vc$ at iteration $t$ is given by 
\begin{align}
	\vc^{(t)} = P_{\cC}\left\{\vc^{(t-1)}+\rho_c\left[\lvert\bTheta\rvert\mathbf{1} + (\mX - \mC^{(t-1)}\mF\rT)\va\right]\right\},\notag
\end{align}
where $P_{\cC}\{\cdot\}$ is defined in \eqref{eq:prox}, $\rho_c$ is the step size, and the gradient is provided in the appendix. 

\paragraph{Updating \mF, given \vc}
For a fixed $\vc$, $(\mathcal{P}2)$ reduces to
\begin{align}
	&\underset{\mF}{\text{maximize}\quad}  
	-\|\mX-\mC\mF\rT\|_F^2 -\alpha \|\mF\|_F^2,\notag
\end{align}
which is an $\ell_2-$regularized linear regression problem. The update for $\mF$ at iteration $t$ using gradient ascent is given by
\begin{align}
	\mF^{(t)} = \mF^{(t-1)}+\rho_{F}\left[(\mX\rT - \mF^{(t-1)}\mC\rT)\mC - \alpha\mF^{(t-1)}\right],\nonumber
\end{align}
where $\rho_F$ is the step size. See the appendix for the gradient.

To summarize, the inference algorithm related to \texttt{GA-Affine-Real} involves two alternating steps of computing $\vc$ and $\mF$ in each iteration of alternating minimization. Updating $\vc$ constitutes of several gradient ascent steps, in which $|\bTheta|\mathbf{1}$, $\mX\va$, and $\mF\rT\va$ cost $N^2$, $ND$, and $2D$ flops, respectively, and can be precomputed. Each iteration of the $\vc$-step in the inference algorithm related to \texttt{GA-Affine-Real}, excluding the precomputations, costs approximately order $N$ flops. Therefore, the $\vc$-step approximately costs $k_1N+N^2$ flops, where $k_1$ is the number of gradient ascent iterations. While updating $\mF$, we can precompute $\mX\rT\mC$ and $\mC\rT\mC$,  which cost $2DN$ and $4N$ flops, respectively. The $\mF$-step, therefore, approximately costs $k_2D+DN$ flops, where $k_2$ is the number of gradient ascent iterations required to update $\mF$. The algorithm, therefore, approximately costs order $k_1N+N^2+k_2D+ND$ flops per one step of the alternating minimization procedure. 

Although the algorithms derived from the \texttt{GA-Affine} models are computationally efficient, we may not always be able to capture the relationship between the attribute of a node and its core score by simple affine functions. Furthermore, we may want to incorporate the information about spatial distances between the nodes, which is impossible using the one-to-one modeling as done by the \texttt{GA-Affine} models. We, therefore, next propose non-linear generative models that address these limitations.     

\section{Nonlinear Models \label{sec:nonlinear models}}
In some cases, correlations between the node attributes are more relevant to the core scores of different nodes than just their respective attributes. For instance, when the node features are spatial positions, the relative distances between two features, rather than the spatial positions themselves, are more relevant to the coreness of nodes. Specifically, two spatially close nodes are more likely to be connected and be in the core part of the network than two spatially well-separated nodes. However, nothing as such can be commented on the coreness of a node based on its spatial position. In what follows, instead of relying on an affine relationship between attributes of nodes to their respective core scores, we propose two models that relate attributes of two nodes to their core scores through nonlinear functions. The two models handle the following two different cases: $(i)$ when both node attributes and graph are observed and $(ii)$ when only node attributes are observed. In the first case, we relate the covariance of attributes of two nodes to the cores scores of the two nodes while employing the probabilistic model for $\bTheta$ from~\eqref{eq:theta;c}, and in the second case, we relate the attributes of nodes to the cores scores through a latent graph.

\subsection{Learning from Node Attributes and Graph}
We first propose a generative model, which models the core score values in terms of partial correlations between the node attributes and spatial distances between the nodes. In a network with a core-periphery structure, typically, attributes of nodes with large core scores should have a high partial correlation as they are strongly connected, the partial correlation between attributes of nodes with large and small core scores should be relatively low, and the partial correlation between attributes of nodes with small core scores should be very low as they are very sparsely connected. Also, for similar reasons, attributes of spatially far apart nodes should have a relatively lesser partial correlation. Inspired by the Gaussian graphical model~\cite{Friedman2008Sparse}, we build a similar model to relate core scores to node attributes, i.e., we model $\vx_k$ for $k=1,\cdots,D$ as $\vx_k\sim\cN(\mathbf{0},\mathbf{K}^{-1}(\vc))$, where $\mathbf{K}(\vc)$ is the precision matrix, which depends on $\vc$. The inverse covariance matrix $\mathbf{K}$ is related to the partial correlation $\rho_{ij}$ between the $i$th and the $j$th variables, $\vx_i$ and $\vx_j$, given other variables as 
%
$$\rho_{ij}= \frac{-K_{ij}}{\sqrt{K_{ii}}\sqrt{K_{jj}}}=\frac{1-w_{ij}}{\sqrt{w_{ii}-1}\sqrt{w_{jj}-1}},$$ where $w_{ij}=1-c_i-c_j+e{\rm log}(d_{ij}+\epsilon)$ and $\epsilon$ is set to a very small value (e.g., $10^{-5}$). Modeling the partial correlations this way results in a high partial correlation between attributes of nodes with large core scores than those with smaller core scores. As the parameter $e$ increases, the relative importance of spatial distances compared to core scores increases. The parameter $e$ can be set to $0$ when the spatial distances are not relevant or unavailable. 
Specifically, we propose the following model for $p(\mX;\vc)$ \begin{align}
	p(\mX;\vc)
	& \propto {\rm det}(\mathbf{K}){\rm exp}(-\tr(\mS\mathbf{K})),
	\label{eq:X|creal-b}
\end{align}
where $\mS = \frac{1}{d} \sum\limits_{i=1}^d\vx_i\vx_i\rT$ is the sample covariance matrix. We refer to this model as \texttt{Graph-Attributes-Nonlinear (GA-Nonlinear)} and is summarized in Fig.~\ref{fig:modelOG}(c). 

\subsubsection*{Maximum Likelihood Estimator} \label{subsec:nonlin mle}
To derive an estimator for inferring core scores from the \texttt{GA-Nonlinear} model, we fit node attributes to the model and estimate $\vc$. Specifically, we maximize the log-likelihood of $\mX$ and $\bTheta$ parameterized by $\vc$. Using the fact that $\mX$ and $\bTheta$ are conditionally independent given $\vc$, we have
\begin{align}
	\mathcal{L}_3(\vc) &=  {\rm log} \, p(\mX,\bTheta;\vc) 
	=  {\rm log} \, p(\mX;\vc)+ {\rm log} \, p(\bTheta;\vc),\notag\\
	&= \log \operatorname{det} \mathbf{K} - \operatorname{tr}(\mathbf{S} \mathbf{K}) + \sum\limits_{i,j=1}^{N} (c_{i}+c_{j})\lvert\Theta_{ij}\rvert\nonumber
\end{align} 
Therefore, the proposed maximum likelihood optimization problem to infer $\vc$ under the \texttt{GA-Nonlinear} model is given by  
\begin{align}
	(\cP3):\,\, &\underset{\vc \in \mathcal{C}}{\text{maximize}\,}  \,\,
	\log \operatorname{det} \mathbf{K} - \operatorname{tr}(\mathbf{S} \mathbf{K}) + \sum\limits_{i,j=1}^{N} (c_{i}+c_{j})\lvert\Theta_{ij}\rvert
	\nonumber \\
	&\text{s. to\quad }  K_{ij} = - c_i - c_j + e\, {\rm log}(d_{ij}),\, i,j=1,\cdots,N,\notag
\end{align}
where $\cC$ is the constraint set defined in \eqref{eq:constraintC}. This is a convex optimization problem in the variable $\vc$ and we solve it using projected gradient ascent. 

The update of $\vc$ at iteration $t$ of projected gradient ascent is given by 
\begin{align}
	\vc^{(t)} = P_{\cC}\left\{\vc^{(t-1)}+\rho_{c}\left[\lvert\bTheta\rvert\mathbf{1}-[{\mathbf{K}}^{-1}(\vc^{(t-1)})]\mathbf{1}+\mS\mathbf{1}\right]\right\},\notag
	\label{eq:gradient ascent real-b}
\end{align}
where $P_{\cC}\{\cdot\}$ is defined in \eqref{eq:prox} and the gradient is provided in the appendix. Here, $\mathbf{K}^{-1}(\vc^{(t)})$ means the inverse of $\mathbf{K}(\vc^{(t)})$. 

Updating $\vc$ involves computing the inverse of an $N$-dimensional matrix. Therefore, the above update approximately costs $k_1N^3$ flops per iteration of projected gradient ascent, where $k_1$ is the number of projected gradient ascent steps. Although it is observed that the inference algorithm related to \texttt{GA-Nonlinear} is computationally more expensive than the inference algorithms related to \texttt{GA-Affine}, it comes with the advantage of being able to incorporate spatial distance information while inferring core scores.

With both node attributes and graph as input, we can use the algorithms derived from the proposed models \texttt{GA-Affine} and \texttt{GA-Nonlinear} to estimate the core scores. However, we may sometimes have access only to node attributes, and the underlying graph may not be available. We next propose a model that captures the dependence of node attributes on core scores through a latent graph so that when only node attributes are available, we can fit data to the model to infer the graph and its core scores jointly.
\subsection{Learning from only Node Attributes}
We model the generative process of $\mX$ from $\bTheta$ by a Gaussian graphical model. Ignoring the normalizing factors, the conditional PDF of $\mX$ given the graph structure $\bTheta$ follows
\begin{equation}
	p(\mX|\bTheta) \propto \operatorname{det}\bTheta\,\,  {\rm exp}(-\operatorname{tr}(\mathbf{S} \bTheta)),
	\label{eq:p(X|theta)}
\end{equation} 
where $\mS$ is the sample covariance matrix. Next, to model the dependence of the latent graph structure on core scores, we employ the probabilistic model for $\bTheta$ from~\eqref{eq:theta;c}, where we model the $(i,j)$th entry of $\bTheta$ as a random variable drawn from a Laplace distribution with inverse diversity parameter $w_{ij}$ as in~\eqref{eq:theta;c}. We refer to this model as \texttt{Attributes-Only(AO)} and it is summarized in Fig.~\ref{fig:modelOG}(d). In what follows, we discuss a learning algorithm for simultaneously inferring $\vc$ and $\bTheta$ by fitting the observed node attributes $\mX$ to this model. 

\subsubsection{Maximum a Posteriori Estimator of $\bTheta$ and $\vc$}  \label{sec:inferringthetaandc}
To jointly estimate $\bTheta$ and $\vc$, we maximize the posterior distribution $p(\bTheta\mid\mX;\vc)$ of $\bTheta$ given node attributes to estimate the model parameters, i.e., we maximize
\[p(\bTheta\mid\mX;\vc)=\frac{p(\mX\mid\bTheta)p(\bTheta;\vc)}{p(\mX)}\] with respect to  $\bTheta$ and $\vc$, where \[p(\mX)=\int_{\bTheta}p(\mX\mid\bTheta)p(\bTheta;\vc)\,d\bTheta.\] Taking logarithm on both sides and ignoring terms independent of the learnable parameters $\bTheta$ and $\vc$, maximizing the posterior distribution is equivalent to maximizing 
\begin{align}
	\cL_4(\bTheta,\vc) &=  {\rm log} \, p(\mX | \bTheta) +{\rm log}\, p(\bTheta;\vc)\notag\\
	&= {\rm log}({\rm det}(\bTheta))-{\rm tr}(\mS\bTheta) + Z -\lambda \sum_{i,j=1}^{N}\left( w_{ij}|\Theta_{ij}|\right) \nonumber
\end{align} 
with respect to $\bTheta$ and $\vc$. Here, the log-likelihood term $ {\rm log} \, p(\mX | \bTheta)$ is from~\eqref{eq:p(X|theta)} and the logarithm of the prior distribution ${\rm log}\, p(\bTheta;\vc)$ is from~\eqref{eq:theta;c} with $Z$ being the normalization constant, which does not depend on $\bTheta$ and $\vc$. Then the proposed optimization problem for jointly estimating $\bTheta$ and $\vc$ is 
\begin{align}
	(\mathcal{P}4):\underset{\bTheta \succ 0, \vc\in \cC}{\text{maximize}\quad}  
	&\log \operatorname{det} \bTheta - \operatorname{tr}(\mathbf{S} \bTheta) - \lambda \sum\limits_{i,j=1}^{N} w_{ij}\lvert\Theta_{ij}\rvert  
	\nonumber \\
	\text{s. to\quad}\, & w_{ij} = 1- c_i - c_j + e\, {\rm log}(d_{ij})\nonumber\\
	& w_{ij} > 0, \quad i,j=1,2,\ldots,N,\,  \notag
	\label{eq:proposed}
\end{align}
where recall the constraint set $\cC$ defined in \eqref{eq:constraintC}. The proposed prior distribution of $\bTheta$ introduces a weighted $\ell_1$ regularization in the optimization problem, where the $\ell_1$ penalty on an edge is determined by the core scores of the constituent nodes of the edge and the spatial distance between the nodes. The penalty is small when both the nodes, as explained by the data, belong to the core part of the network and are spatially close. As the parameter $e$ increases, the percentage of edges between spatially distant nodes decreases. However, $e$ should be chosen while ensuring that $w_{ij}>0$, which is necessary for the prior on $\bTheta$ to be a valid distribution.

The proposed problem $(\mathcal{P}4)$ is a non-convex optimization problem in the variables $\bTheta$ and $\vc$. We present an iterative algorithm to maximize the objective function in $(\mathcal{P}4)$ alternatingly, wherein we update $\bTheta$ while fixing $\vc$ and update $\vc$ while fixing $\bTheta$ till convergence. Each of the two sub-problems in the alternating steps of the algorithm for updating $\vc$ and $\bTheta$ are convex optimization problems.

\paragraph{Updating $\bTheta$, given $\vc$}
For a fixed $\vc$, the $(\cP4)$ simplifies to the following convex optimization problem
\begin{equation}
	\underset{\bTheta \succ 0}{\text{maximize}\quad} \log \operatorname{det} \bTheta-\operatorname{tr}(\mathbf{S} \bTheta)-\lambda \sum_{i,j=1}^{N}w_{ij}\lvert \Theta_{ij}\rvert,\nonumber
	\label{eq:theta step}
\end{equation}
where the weights of the $\ell_1$ regularization term depend on the core scores, which, unlike in $(\mathcal{P}4)$, are now known. The sparsity pattern in $\bTheta$ is influenced by the core scores estimated in the previous iteration of the alternating minimization and the data $\mX$. The problem can be solved using existing efficient solvers such as QUIC~\cite{Hsieh2014QUIC}.

\paragraph{Updating $\vc$, given $\bTheta$}
Fixing $\bTheta$, $(\mathcal{P}4)$ simplifies to the following linear program 
\begin{align}
	&\underset{\vc \in \mathcal{C}}{\text{maximize}\;\;} 
	\sum_{i,j=1}^{N}|\Theta_{ij}|(c_i+c_j) \nonumber\\
	&\text{s. to \quad}\,
	c_i + c_j < 1 + e {\rm log} (d_{ij}),\,\, i,j= 1,\cdots,N.
	\label{eq:c step}
\end{align}
The problem can be solved using standard linear program solvers. The update of core scores depends on the previous estimate of $\bTheta$. The sub-problem in~\eqref{eq:c step} can itself be seen as another core score inference technique that learns core scores from a connectivity structure as in the existing works discussed in Section~\ref{subsec:PriorWorks}. 

To summarize, the inference algorithm related to \texttt{AO} involves two alternating steps of computing $\bTheta$ and $\vc$ in each iteration of alternating minimization. Updating $\bTheta$ using QUIC constitutes of several coordinate ascent steps, each of which costs $N$ flops~\cite{Hsieh2014QUIC}. Therefore, the $\bTheta$-step approximately costs $k_1N$ flops, where $k_1$ is the number of coordinate ascent iterations. Updating $\vc$ using standard linear program solvers approximately costs $N^{3.5}B$ flops, where $B$ is the number of bits in the input~\cite{Karmarkar1984A}. The algorithm, therefore, approximately costs order $k_1N+BN^{3.5}$ flops per one step of the alternating minimization algorithm. Although the algorithm is computationally more expensive than the inference algorithms related to \texttt{GA-Affine} and \texttt{GA-Nonlinear}, it is due to the cost involved in inferring both the graph and core scores, where, in contrast, only the core scores are inferred in the algorithms related to the other models discussed in the paper.  
\begin{remark}
Problem $(\cP4)$ reduces to the classical graphical lasso~\cite{Friedman2008Sparse} by fixing $w_{ij}=1$ for $i,j=1,\cdots,N$. In graphical lasso, the $\ell_1$-penalty is uniformly applied on all the entries of $\bTheta$. Therefore, no specific sparsity structure, such as the core-periphery structure of interest, is incorporated in it. 
\end{remark}

\section{Numerical Experiments \label{sec:NumericalExp}}
In this section, we evaluate the proposed models and algorithms in terms of their ability to identify core and periphery parts of several synthetically generated and real-world networks. 
\subsection{Dataset Description \label{subsec:DataDescription}}
We begin by describing all the synthetic and real-world datasets that we use in the remainder of the section. 
\subsubsection{Synthetic Datasets}
We synthetically generate data according to the four generative models proposed in this work.


\paragraph{\texttt{Graph-Attributes}}
For generating data according to the \texttt{GA} models, we start with a core score vector $\vc$ and spatial distances $d_{ij}$ between nodes $i$ and $j$ for $i,j=1,\cdots,N$. Specifically, we consider a graph with $60$ nodes. The first $a\%$ entries of $\vc$ of those nodes considered to belong to the core part of the network are set uniformly at random to values between $0.9$ and $1$. The rest of the entries correspond to the periphery nodes and are set uniformly at random to values between $0$ and $0.01$. The logarithm of the distances between the core nodes are set uniformly at random to values between $1$ and $1.05$ and the logarithm of the distances between the peripheral nodes are set uniformly at random to values between $1.2$ and $1.205$. The parameter $e$ is set to $1$. We then generate the edges connecting nodes $i$ and $j$, for $i,j=1,\cdots,N$, of the underlying graph according to~\eqref{eq:theta;c}.
We generate graphs with different percentages of nodes in the core part. In particular, we generate data for $a \in \{10,50,90\}$. We then generate node attributes from $\vc$ according to the \texttt{GA-Affine-Bool}, \texttt{GA-Affine-Real}, and \texttt{GA-Nonlinear} models using  \eqref{eq:X|cbinary}, \eqref{eq:X|creal-a}, and \eqref{eq:X|creal-b}, respectively. To generate node attributes according to the \texttt{GA-Affine-Real} and \texttt{GA-Affine-Bool} models, we choose a matrix with random entries drawn from a standard normal distribution as $\mF\in \mathbb{R}^{D\times2}$. 

\paragraph{\texttt{Attributes-Only}}
Next, to generate data according to the \texttt{AO} model, the generation process of $\bTheta$ from $\vc$ remains the same. In fact, we use the same graphs as earlier for all the experiments. However, the node attributes are now generated using $\bTheta$. We draw $D=30$ samples from a multivariate Gaussian distribution with mean $\mathbf{0} \in \mathbb{R}^{N}$ and precision matrix $\bTheta \in \mathbb{R}^{N\times N}$ to generate the columns of node attribute matrix $\mX\in \mathbb{R}^{N\times D}$ according to~\eqref{eq:p(X|theta)}. 

\subsubsection{Real-world datasets}
We employ $8$ datasets with real-valued node attributes and $4$ datasets with binary-valued node attributes, for all of which both graph and node attributes are available. The datasets with real-valued node attributes that we use in our experiments are \emph{C. elegans}~\cite{Sen2008Collective}, \emph{London underground}~\cite{Jia2019Random}, \emph{Twitter (Olympics)}~\cite{Greene2013Producing}, \emph{Freeman}~\cite{Freeman1979the}, \emph{Organizational (Advice)}, \emph{Organizational (Value)}, \emph{Organizational (R\&D Advice)}, \emph{Organizational (R\&D Aware)}~\cite{Cross2004hidden}, and \emph{Openflights}~\cite{Jia2019Random}. The datasets with binary-valued node attributes that we use in our experiments are \emph{Cora}~\cite{Sen2008Collective}, \emph{Facebook}~\cite{leskovec2012Learning}, \emph{Twitter}~\cite{leskovec2012Learning}, and \emph{Google plus}~\cite{leskovec2012Learning}. 
We also use a brain network dataset from the \emph{OHSU} brain institute [26] to test the usefulness of the inference algorithm related to \texttt{AO} in performing a downstream machine learning task, namely, classification; see Table~I for more details on the datasets. 
\begin{table}
	\centering
	\resizebox{\columnwidth}{!}{
	\begin{tabular}[width = \columnwidth]{c c c c c}
		\hline
		{Datasets} & {$N$} & {$D$} & {Real/Boolean} & {Spatial distances}   \\ \hline
		\emph{C. elegans} & 131 & 2 & Real & Available \\ 
		London underground & 303 & 2  & Real & Available \\ 
		Twitter (Olympics) & 464 & 3097  & Real & Not relevant\\ 
		Freeman & 32 & 3 & Real & Not relevant \\ 
		Organizational (Advice) & 46 & 4 & Real & Not relevant \\ 
		Organizational (Value) & 46 & 4 &  Real & Not relevant \\ 
		Organizational (R\&D Advice) & 77 & 3  & Real & Not relevant \\ 
		Organizational (R\&D Aware) & 77 & 3 &  Real & Not relevant \\ 
		Openflights & 7184 & 1  & Real & Available \\
		Cora     & 2708 & 1433 & Boolean & Not relevant \\ 
		Facebook    & 347 & 224 & Boolean & Not relevant   \\ 
		Twitter   & 244 & 1364  & Boolean & Not relevant  \\ 
		G+ & 1692 & 1319 & Boolean & Not relevant  \\  
		OHSU & 190 & 74  & Real &  Available  \\ \hline
		\label{table:datasets}
	\end{tabular}
	}
	\newline\newline
	\vspace{-4mm}
	\caption{Details of datasets.}
	\vspace{-4mm}
\end{table}
\subsection{Baselines and Metrics \label{subsec:BaselinesforComparison}}
We compare the core scores learnt using our methods with existing methods for learning core scores described in Section~\ref{subsec:PriorWorks}. Specifically, we compare our algorithms with \texttt{MINRES}~\cite{Boyd2010Computing}, \texttt{Rombach}~\cite{Rombach2014Core}, \texttt{Random-Walk} \cite{Della2013Profiling}, \texttt{k-cores}~\cite{Della2013Profiling}, all of which take only the graph as input and ignore the node attributes. We also compare the graph learning ability of the proposed algorithm with \texttt{AO}
and \texttt{Graphical Lasso}~\cite{Friedman2008Sparse}. 
\begin{table*}
	\centering
	\resizebox{2\columnwidth}{!}{%
		\begin{tabular}[width = 0.5\columnwidth]{c c c c c c c c c}
			\hline
			a & \texttt{GA-Affine-Real} & \texttt{GA-Nonlinear} & \texttt{GA-Affine-Bool}  & \texttt{AO} & \texttt{MINRES}    & \texttt{Rombach} & \texttt{Random-Walk}   & \texttt{k-cores}   \\ \hline
			$10\%$   & $\mathbf{0.9995\pm1.2\times10^{-4}}$ & $0.9993\pm8.0\times10^{-4}$ & $\mathbf{0.9995\pm6.7\times10^{-5}}$ & $0.570\pm0.236$ & $0.418\pm0.054$ & $0.742\pm0.003$ & $0.437\pm0.042$ & $0.662\pm0.071$ \\
			$50\%$   & $\mathbf{0.9995\pm5.9\times10^{-5}}$  & $\mathbf{0.9995\pm5.9\times10^{-5}}$ & $\mathbf{0.9995\pm5.9\times10^{-5}}$ & $0.857\pm0.042$& $0.903\pm0.008$  & $0.806\pm0.003$ & $0.856\pm0.005$ & $0.981\pm0.004$ \\ 
			$90\%$   & $\mathbf{0.9995\pm4.8\times10^{-5}}$  & $\mathbf{0.9995\pm4.8\times10^{-5}}$ & $\mathbf{0.9995\pm4.8\times10^{-5}}$ & $0.952\pm0.018$& $0.938\pm0.006$  & $0.699\pm0.003$ & $0.853\pm0.004$ & $0.997\pm0.006$ \\ \hline
		\end{tabular}%
	}
	\newline\newline
	\caption{Cosine similarity scores between the groundtruth core score vectors and the estimated core score vectors.}
	\label{table:syn S_c}
	\vspace{-4mm}
\end{table*}

\begin{table}
	\centering
	\resizebox{\columnwidth}{!}{
	\begin{tabular}{c c c c c c c}
		\hline
		& \texttt{GA-Affine-Real}  & \texttt{AO} & \texttt{MINRES}    & \texttt{Rombach} & \texttt{Random-Walk}   & \texttt{k-cores}   \\ \hline		
		Organizational (Advice)    & 14.558  & 17.448 & 15.777   & 12.121 & 15.490 & 14.506 \\ 
		Organizational (Value)    & 21.487  & 24.046 & 23.101   & 20.116 & 21.519 & 21.556 \\ 
		Organizational (R\&D Advice) & 26.018    & 28.707  & 27.476  & 25.046 & 26.805 & 23.684 \\ 
		Organizational (R\&D Aware) & 34.799  & 39.154  & 38.852  & 33.224 & 36.445 & 33.933 \\ 
		Twitter (Olympics)    & 131.605  & 134.692 & 137.142   & 124.112 & 131.278 & 129.221 \\ 
		Freeman & 7.322 & 7.801  & 8.114  & 6.585 & 7.647 & 7.599 \\ 
		{\emph{C. elegans}}    & {40.460} & {41.94} & {41.82}  & {39.076} & {40.877} & {39.051} \\
		{London underground}   & {79.988} & {79.216} &  {79.249} & {78.7563}  & {79.338} & {79.169}  \\ 
		& \texttt{GA-Nonlinear} & \texttt{AO} & \texttt{MINRES}    & \texttt{Rombach} & \texttt{Random-Walk}   & \texttt{k-cores}   \\ 
		\emph{C. elegans}    & 40.311 & 41.94 & 41.82  & 39.076 & 40.877 & 39.051 \\
		London underground   & 79.263 & 79.216 &  79.249 & 78.7563  & 79.338 & 79.169  \\ 
		Twitter (Olympics)   & 134.796 & 134.692 & 137.142   & 124.112 & 131.278 & 129.221 \\ \hline
	\end{tabular}
	}
	\newline\newline
	\caption{$\|\bTheta_{\rm ideal}-\bTheta_0\|_F$ for different core score estimation algorithms on datasets with real-valued node attributes.}
	\label{table:res_real}
\end{table}

\begin{table}
	\centering
	\resizebox{1\columnwidth}{!}{
	\begin{tabular}{c c c c c c}
		\hline
		& \texttt{GA-Affine-Bool} & \texttt{MINRES}    & \texttt{Rombach} & \texttt{Random-Walk}   & \texttt{k-cores}   \\ \hline
		Cora     & 678.991   & 680.503 & 678.286 & 679.906  & 678.389 \\ 
		Facebook    & 89.721  & 106.371 & 87.835 & 98.422  & 88.335 \\ 
		Twitter   & 77.285  & 75.598 & 67.283 & 70.534   & 66.656 \\ 
		G+ & 543.598 & 513.716 & 473.931  & 499.452  & 466.918  \\ \hline
	\end{tabular}
	}
	\newline\newline
	\caption{$\|\bTheta_{\rm ideal}-\bTheta_0\|_F$ for different core score estimation algorithms on datasets with binary-valued node attributes.}
	\label{table:res_binary}
	\vspace{-4mm}
\end{table}

To evaluate the efficacy of our algorithms, we estimate the core scores for synthetically generated datasets and compare them with the groundtruth core score vectors. In particular, we compute the cosine similarity score between the actual and the predicted core score vectors from different core score learning algorithms, where the cosine similarity metric $\cS(\va,\vb)$ between two vectors $\va$ and $\vb$ is defined as
$$
\cS(\va,\vb) = \frac{\va\rT\vb}{\|\va\|_2\|\vb\|_2}.
$$
A higher cosine similarity indicates a better agreement between the actual core score vector and the estimated one. 

For real-world datasets, when the groundtruth core score vectors are not available, we follow the following approach. We permute the rows and columns of groundtruth adjacency matrices in the decreasing order of the core scores output by different methods and normalize it so that its entries lie between $0$ and $1$. We denote the normalized permuted adjacency matrices by $\bTheta_0$. We then compute the Frobenius norm of the difference between the ideal core-periphery model~\cite{Borgatti2000Models}
\begin{equation}
	\boldsymbol{\Theta}_{\text {ideal }}=\left[\begin{array}{l|l}
		\mathbf{1}_{a a} & \mathbf{0}_{a(N-a)} \\
		\hline \mathbf{0}_{(N-a) a} & \mathbf{0}_{(N-a)(N-a)}
	\end{array}\right]
	\label{eq:idealcp}
\end{equation}
and the normalized permuted adjacency matrices of different methods. In all our experiments, we choose $a=N/4$, where $N$ is the number of nodes.

To evaluate the ability of the inference algorithm related to \texttt{AO} to correctly learn the structure of a graph while simultaneously estimating the core scores, we compare its performance with that of \texttt{Graphical Lasso}. We quantify the closeness of estimated graphs using the two methods to the groundtruth graph by computing the cosine similarity between the vectorized versions of the adjacency matrix of the actual and estimated graphs.

\begin{table*}
\centering
	\hspace*{16mm}
	\begin{tabular}{c c c c}
		\hline
		& $10\%$ core & $50\%$ core  & $90\%$ core  \\ \hline
		\texttt{AO}  & $0.157\pm0.056$  & $0.359\pm0.035$ & $0.40\pm0.038$\\ 
		\texttt{Graphical lasso}   & $0.147\pm0.024$ & $0.233\pm0.025$ & $0.261\pm0.033$ \\ \hline
	\end{tabular}
	\newline\newline
	\caption{Comparison of the inference algorithm related to \texttt{AO} with graphical lasso on synthetically generated data.}
	\label{table:glasso_comp}
\end{table*}

\begin{figure*}[t]
	\centering
	\includegraphics[width=2\columnwidth]{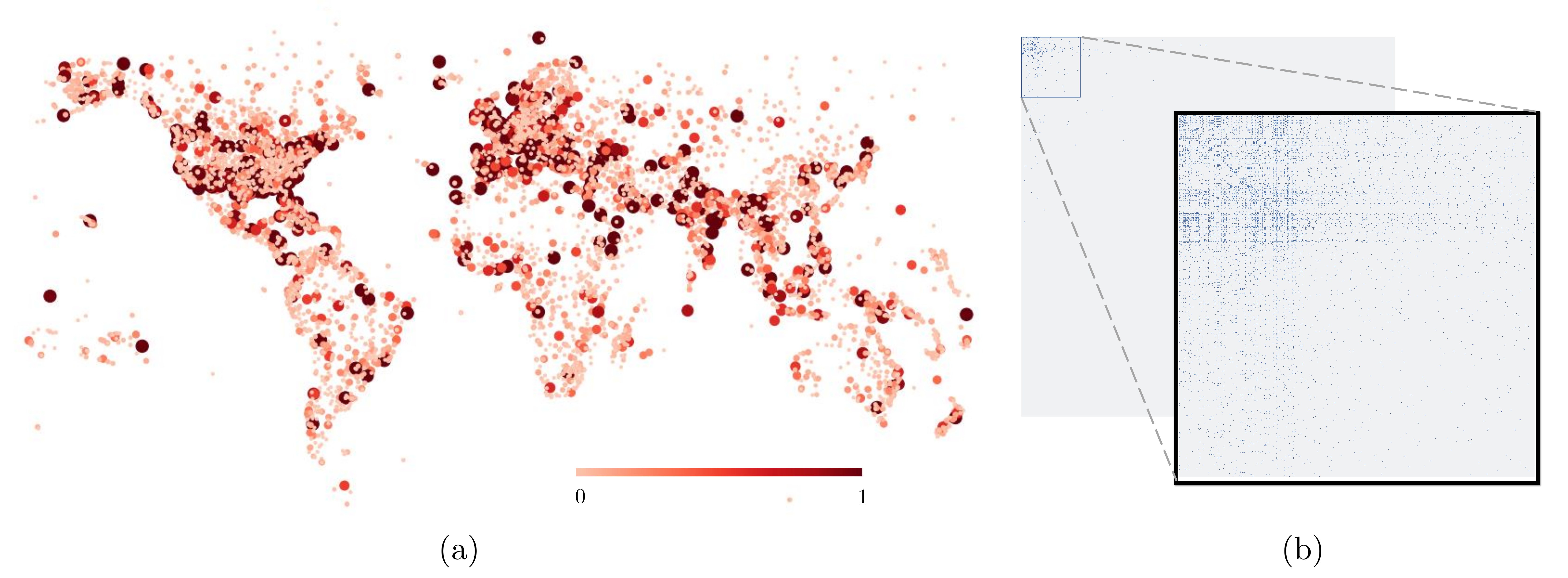}
	\caption{(a) Core scores of different airports and (b) ground truth adjacency matrix of airports network with rows and columns permuted according to the decreasing order of core scores.
	}
	\label{fig:openflights}
\end{figure*}
\subsection{Learning Core Scores \label{seubsec:InferringcoreScores}}
We take $20$ different core score vectors each with $10\%, 50\%$, and $90\%$ of the nodes in the core part and then generate graphs and node attribute matrices from them using the proposed models. 
We provide the graphs as input to the existing methods to estimate the core scores. Out of the proposed algorithms, the \texttt{GA-Affine-Bool}, \texttt{GA-Affine-Real}, and \texttt{GA-Nonlinear} algorithms take both the node attribute matrix generated by the corresponding model and the graph structure while the inference algorithm related to \texttt{AO} takes only the node attributes as input to estimate the core scores. The cosine similarity scores between the estimated core score vectors from each of these methods and the groundtruth core score vector averaged over the $20$ datasets for different percentages of core part are shown in Table~\ref{table:syn S_c}. The values that follow $\pm$ in each of the entries in the table correspond to the standard deviations of the cosine similarity values obtained for the $20$ datasets. The merit of the proposed algorithms in correctly estimating the core scores by capturing information from both node attributes and graph in comparison to the existing methods that take only the graph as input is clearly evident from the highest cosine similarity values of the inference algorithms related to \texttt{GA}. Although the inference algorithm related to \texttt{AO} takes only the node attribute matrix, it can be seen to have comparable performance as the other existing methods.

We also test the proposed algorithms on real-world data, both with real and binary-valued node attributes. In datasets with real-valued node attributes, if the core score of a node is influenced by only the attribute of that node, we learn the core scores using the inference algorithm related to \texttt{GA-Affine-Real}. If the dependencies between attributes of all the adjacent nodes of a node influence its core score, we use the inference algorithm related to \texttt{GA-Nonlinear} to infer the core scores. For example, we infer the core scores of the Freeman dataset using the inference algorithm with \texttt{GA-Affine-Real} as the number of citations of a researcher directly influences the coreness of that researcher. In datasets like \emph{C. elegans} and London underground, where the node attributes are spatial positions, the spatial distances of a node from all its neighboring nodes influence the coreness of the node. Therefore, for such datasets, we use the inference algorithm related to \texttt{GA-Nonlinear}. It can be observed from Table~\ref{table:res_real} and Table~\ref{table:res_binary} that for all the datasets, the quantity $\|\bTheta_{\rm ideal}-\bTheta_0\|_F$ obtained for different algorithms are comparable. This indicates that the proposed methods correctly identify the cohesively connected nodes in different networks. To prove the advantage that the proposed inference algorithm based on \texttt{GA-Nonlinear} has when compared to that based on \texttt{GA-Affine-Real} in datasets where the dependencies between node attributes influence the core scores of nodes, we also test the inference algorithms with \texttt{GA-Affine-Real} on the \emph{C. elegans} and the London underground datasets. As expected, the values of $\|\bTheta_{\rm ideal}-\bTheta_0\|_F$ obtained with the algorithm based on \texttt{GA-Affine-Real} are higher than those obtained with the algorithm based on \texttt{GA-Nonlinear} [cf. Table~\ref{table:res_real}]. 
It can be observed that even the inference algorithm related to \texttt{AO} that takes only the node attributes as input gives similar performance as that of the existing methods, which take the structure of the graph as input. This indicates that the structural information in the node attributes is correctly captured by \texttt{AO}.   

We next provide a visual representation of the core scores learnt for the Openflights dataset. For this dataset, the node attribute is the degree information. Therefore, the node attribute values directly influence the core scores and we therefore use \texttt{GA-Affine-Real}. The map of airports across the world, where the color and sizes of nodes denote the core scores, is shown in Fig.~\ref{fig:openflights}(a). Here, the darker and bigger nodes denote airports with high core scores and the lighter and smaller nodes denote airports with small core scores.  The top $10$ busy airports in the world (data from Airports Council International), namely, Guangzhou Baiyun,  Hartsfield–Jackson Atlanta, Chengdu Shuangliu, Dallas/Fort Worth, Shenzhen Bao'an, Chongqing Jiangbei, Beijing Capital, Denver, Kunming Changshui, and Shanghai Hongqiao international airports, have been rightly assigned high cores by our method. We can also notice from Fig.~\ref{fig:openflights}(b) that when the rows and columns of the adjacency matrix of the groundtruth network are ordered in the decreasing order of the estimated core scores, a clear core-periphery structure is observed. The submatrix containing the first few rows and columns of the ordered adjacency matrix (containing the core-core connections) is zoomed in for better comprehensibility.
\subsection{Learning a Graph with Core-Periphery Structure \label{subsec:EstimatingGraph}}
We compare the graph reproducibility performance of the inference algorithm related to \texttt{AO} and the classical graphical lasso on the synthetic data generated according to the \texttt{AO} model for different proportions of core nodes as described in Section~\ref{subsec:DataDescription}. We compare cosine similarity between the vectorized forms of the groundtruth adjacency matrix and the predicted adjacency matrices from each of these methods. The values are tabulated in Table~\ref{table:glasso_comp}. We observe that the inference algorithm related to \texttt{AO} outperforms \texttt{Graphical lasso} on all the datasets generated by appropriately capturing the core-periphery structure from the data.

\begin{figure*}
	\centering
	\includegraphics[width=2\columnwidth]{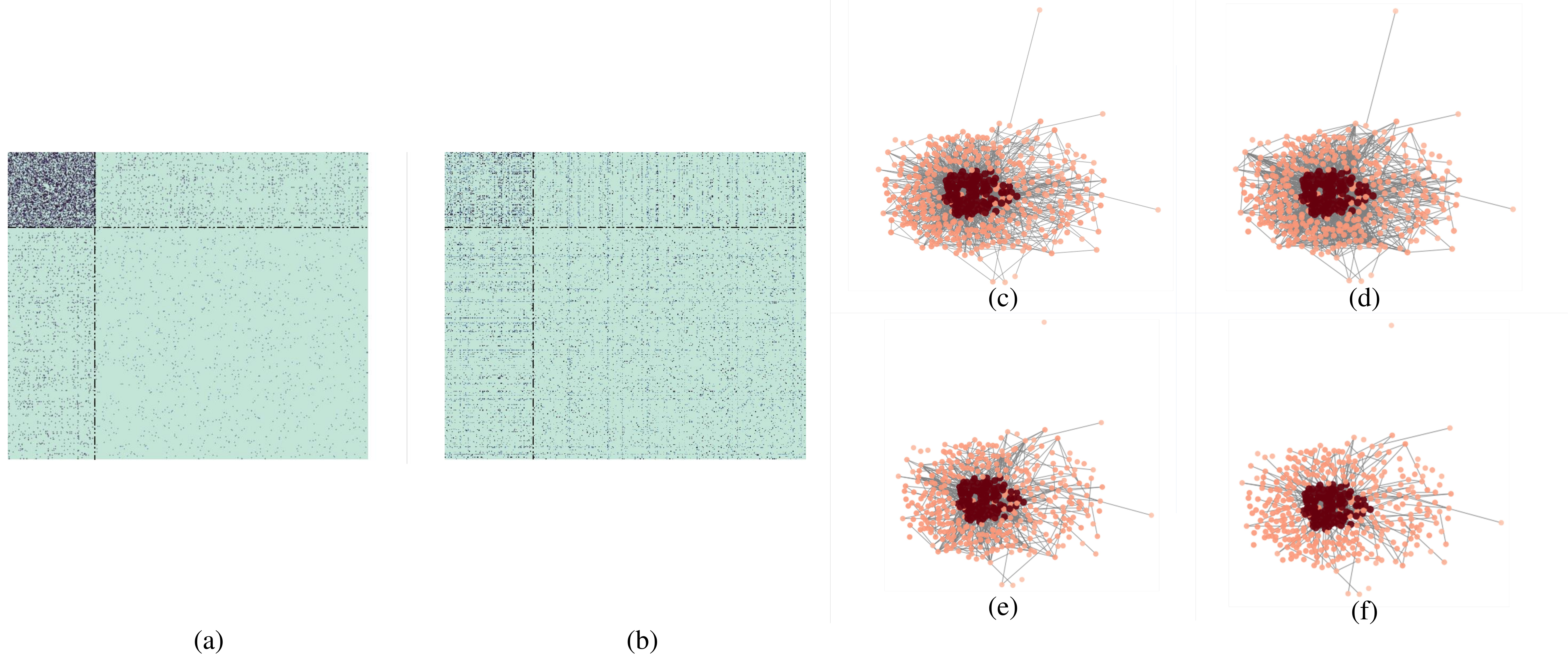}
	\caption{(a) Estimated and (b) ground truth networks of the Twitter dataset ordered in the descending order of the core scores estimated from node attributes. (c) Ground truth graph for Twitter (Olympics) dataset. Graph estimated for Twitter (Olympics) data by the proposed method with (d) $\lambda=0.05$, (e) $\lambda=0.1$, and (f) $\lambda=0.5$.
	}
	\label{fig:Adj twitter}
\end{figure*}

In Fig.~\ref{fig:Adj twitter}, we illustrate the adjacency matrices of the graph estimated by the inference algorithm related to \texttt{AO} and the groundtruth graph for the Twitter (Olympics) dataset. Here, the nodes are ordered in the decreasing order of the core scores predicted by the proposed method. In Fig.~\ref{fig:Adj twitter}(b), we notice that when the rows and columns of the groundtruth graph are arranged according to the core scores, the underlying core-periphery structure is revealed. This validates the correctness of the core scores learnt by the inference algorithm related to \texttt{AO}. Fig.~\ref{fig:Adj twitter}(a) shows that the graph estimated by the proposed method correctly captures the core-periphery structure from the node attributes.
We also illustrate the impact $\lambda$ has on the sparsity structure of the graph estimated. Fig.~\ref{fig:Adj twitter}(c) shows the groundtruth graph for Twitter (Olympics) dataset and Figs.~\ref{fig:Adj twitter}(d)-(f) show graphs estimated with different values of~$\lambda$. The darker nodes are the core nodes as predicted by the proposed inference algorithm, the connections between which form the top left corner of adjacency matrices in Fig.~\ref{fig:Adj twitter}, and the lighter nodes form the periphery. As we increase $\lambda$, the sparsity of the graph increases, as expected. Furthermore, it can be observed that the edges that drop out as $\lambda$ increases are mostly from the periphery region. From  Figs.~\ref{fig:Adj twitter}(d)-(f), we can notice a clear increase in sparsity in the periphery region while the core nodes continue to be cohesively connected.


\subsection{Graph Classification and Analysis with the Brain Dataset\label{subsec:BrainNetworkAnalysis}}

\begin{figure}
	\centering
	\includegraphics[width=\columnwidth]{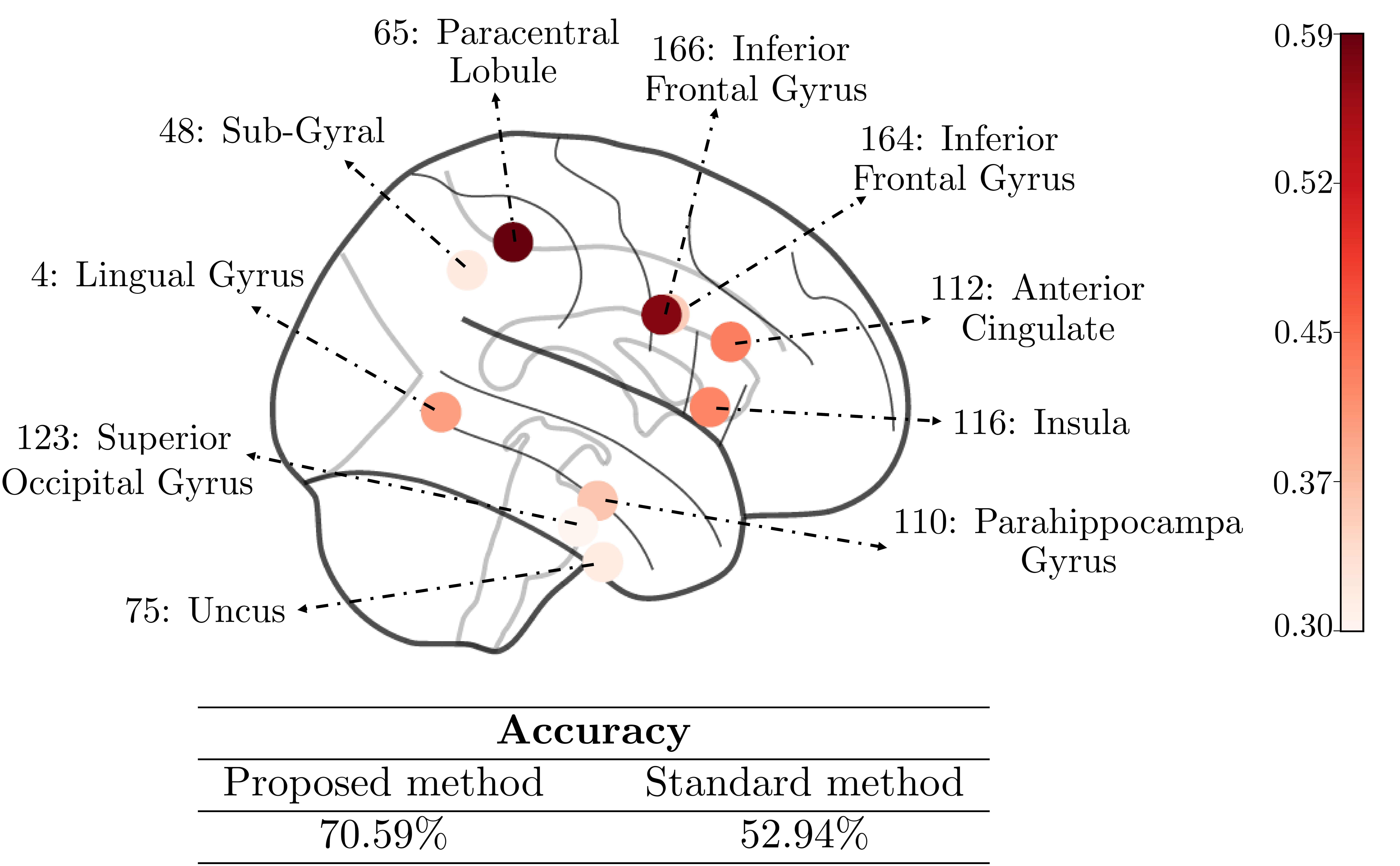}
	\caption{Top: Regions with significant differences in the core scores of healthy individuals and subjects with ADHD. Bottom: Classification accuracy of the GNN classifier trained with the estimated core-periphery graph and the core scores as nodal attributes is significantly higher than the GNN classifier trained with the correlation matrix constructed from the fMRI time series. 
	}
	\label{fig:brain}
\end{figure}
We next perform a downstream machine learning task using the core scores and graphs learnt using the inference algorithm related to \texttt{AO}. Specifically, we classify healthy individuals and subjects with ADHD from their fMRI data in the OHSU dataset using graph neural networks (GNNs)~\cite{Wu2020A}. The OHSU dataset consists of fMRI time series for the regions of interest in the cc200 parcellation for 79 individuals, out of which  $37$ correspond to subjects with ADHD and the other $42$ correspond to healthy subjects. 
We randomly sample about $80\%$ data from the two classes as training data and use the rest for testing.  

We consider a 2-layer GraphSAGE~\cite{Hamilton20017Inductive} model with hidden and output dimensions as 32. The output from the GNN model is given as input to a multi-layer perceptron (MLP) with $2$ layers, which acts as a decoder that takes in the $32$-dimensional representation output by the GraphSAGE model and decodes the class information from it. We demonstrate the merit of the core scores and the graph structure learnt using the proposed inference algorithm by considering two different cases. In the first case, which we refer to as Proposed method, we first compute the core scores and the underlying graphs from fMRI data of different individuals independently using the inference algorithm with \texttt{AO}. We then provide the GNN model with the estimated graph structure $\bTheta$ learnt with \texttt{AO} as the required input adjacency matrix and the learnt core scores as the input node attributes. In the second case, which we refer to as Standard method, we provide the GNN model with the empirical Pearson correlation matrices of the fMRI data as the required input adjacency matrix and an all-one vector as the input node attributes. The accuracies achieved in the two cases are shown in Fig.~\ref{fig:brain}. The accuracy achieved with the Proposed method is significantly higher than that achieved with the Standard method. This increment in performance suggests that the coreness information in the fMRI data is crucial in classifying healthy subjects and subjects with ADHD.
\begin{figure*}[t]
	\centering
	\includegraphics[width=2\columnwidth]{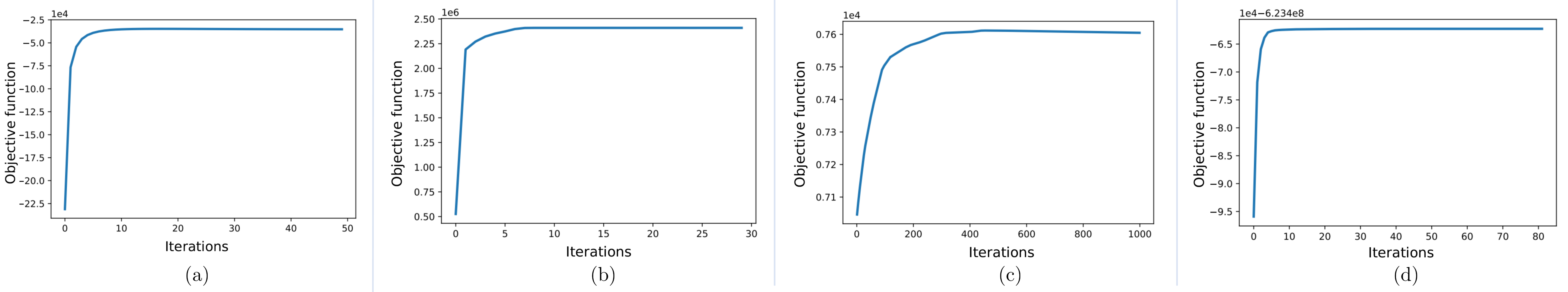}
	\caption{Convergence. (a) \texttt{GA-Affine-Bool} on the Twitter dataset. (b) \texttt{GA-Affine-Real} on the Organizational (R\&D Aware) dataset.  (c) \texttt{GA-Nonlinear} on the London underground dataset. (d) \texttt{AO} on the \emph{C. elegans} dataset. 
	}
	\label{fig:convergence}
\end{figure*}

We also qualitatively visualize the differences in the core and periphery parts of healthy subjects and subjects with ADHD. To do so, we compute the average of the core score vectors from the two classes. The average of the core score
vectors of healthy subjects is denoted by $\bar{\vc}_{\rm HC}$ and that of subjects with ADHD
by $\bar{\vc}_{\rm ADHD}$. To quantify the difference in coreness of different brain regions across the two groups, we compute 
the magnitude of the difference between the average core score vectors of the two classes, i.e., $\vq = |\bar{\vc}_{\rm HC} - \bar{\vc}_{\rm ADHD}|$. The $i$th entry of $\vq$, which quantifies the difference in coreness values of the $i$th brain region across the two groups, implicitly quantifies the difference in connectivity (or interaction) of that region with the others for the two groups. Fig.~\ref{fig:brain} shows the top $10$ regions with the largest difference in connectivity, as measured by the $10$ largest values of $\vq$. The darker nodes in the figure denote the regions with a larger difference in the cores scores of the two groups. The regions with the largest differences in activation for healthy individuals and patients with ADHD as reported by~\cite{Dickstein2014neural} coincide with the regions identified by our method, namely, paracentral lobule, inferior frontal gyrus, anterior cingulate, and insula.
\subsection{Convergence\label{subsec:convergence and run-time}}

In Fig.~\ref{fig:convergence}, we show convergence for each of the four algorithms on four different datasets.
We observe that the inference algorithms based on \texttt{GA-Affine-Bool}, \texttt{GA-Affine-Real}, and \texttt{AO} models converge in less than $50$ alternating minimization iterations when the \emph{tolerance} of the algorithms (and of the sub-problems involved) is fixed to $10^{-4}$, where we define tolerance as the magnitude of difference between the values of objective functions in two consecutive iterations. The method with \texttt{GA-Nonlinear} model does not involve any alternate minimization procedure and involves only gradient ascent steps. We observe that \texttt{GA-Nonlinear} converges in about a few hundred gradient ascent steps when the tolerance is fixed to $10^{-4}$.

\section{Conclusions} \label{sec:conclusions}
We considered on a particular class of graphs known as the core-periphery structured graphs, which are graphs with a group of densely connected core nodes and a group of sparsely connected peripheral nodes. The coreness of a node is specified by a quantity referred to as the core score, which is a scalar representation of a node that quantifies the likelihood of it belonging to the core part of the graph. We proposed probabilistic generative models to relate node attributes and graphs to core scores and derived algorithms from the developed models to infer core scores given a graph and/or its node attributes. Specifically, we proposed two classes of probabilistic graphical models, namely, affine and nonlinear models, to relate node attributes to core scores, which are also used to model the graph structure. The proposed nonlinear generative models for node attributes can further be classified into two classes based on whether they treat the graph as a known or an unknown variable. When the graph is treated as a known variable, we modeled the dependence of node attributes directly on the core scores and developed efficient algorithms to learn the core scores and model parameters. When the graph is treated as an unknown variable, we modeled the dependence of node attributes on the core scores through the latent graph and  presented a joint estimator to infer the core scores and a core-periphery structured graph simultaneously. We observed through experiments on synthetic datasets that the core scores estimated by the proposed algorithms are closer (in the cosine similarity sense) to the actual core scores as compared to the core scores estimated by the existing methods that consider only graphs to infer scores. We also observed that the proposed method that takes only the node attributes as input learns core scores on par with methods that use the groundtruth network as input.

\section*{Appendix}
\renewcommand{\thesubsection}{\Alph{subsection}}

In this section, we provide the expressions for the gradients 
involved in inferring the model parameters of \texttt{GA-Affine-Bool}, \texttt{GA-Affine-Real}, and \texttt{GA-Nonlinear}.

\subsection{\texttt{GA-Affine-Bool}}
The objective function in $(\cP1)$ can be written as
\begin{align}
    \mathcal{L}_1(\vc,\mF)
    =&\sum\limits_{j=1}^{N} 2\vc\rT|\bTheta_{:j}|   -\alpha\|\mF\|_F^2\nonumber\\
    &+ \underbrace{\sum_{i=1}^{N}\sum_{k=1}^{D}{x_{ik}}{\rm log}(\pi_{ik})
	+ (1-x_{ik}){\rm log}(1-\pi_{ik})}_{:=\mathcal{T}_1}  \nonumber,
\end{align}
where $\sum_{j=1}^{N} 2\vc\rT|\bTheta_{:j}| = \sum_{i,j=1}^{N} (c_i +c_j) |\Theta_{ij}|$ with $|\bTheta_{:j}|= [|\Theta_{1j}|,|\Theta_{2j}|,\cdots,|\Theta_{Nj}|]\rT$. We have
\begin{equation}
   {\sum\limits_{j=1}^{N} 2\frac{\partial\vc\rT|\bTheta_{:j}|}{\partial \vc}} = 2\lvert\bTheta\rvert\mathbf{1} 
   \label{eq:gradient_priorc}
\end{equation}

and
\begin{align*}
 	\frac{\partial\mathcal{T}_1}{\partial c_j} &= \sum_{i=1}^{N}\sum_{k=1}^{D}\left({\frac{x_{ik}}{\pi_{ik}} - \frac{1-x_{ik}}{1-\pi_{ik}}}\right) \frac{\partial \pi_{ik}}{\partial c_j}, \nonumber \\
 	&= \sum_{k=1}^{D}(x_{jk} - \pi_{jk})a_k. \nonumber
 	\label{eq:derivative binary 1}
 \end{align*}
as 
 \begin{align}
 \frac{\partial \pi_{ik}}{\partial c_{j}} 
&= \begin{cases}
			a_k\pi_{ik}(1-\pi_{ik}), & \text{if $j=i$,}\\
            0, & \text{otherwise.}
		 \end{cases} \nonumber
 \end{align}
 Then \[\frac{\partial\mathcal{T}_1}{\partial \vc} = (\mX - \mP)\va\] and
 \[
 \frac{\partial\mathcal{L}_1}{\partial \vc} = 2\lvert\bTheta\rvert\mathbf{1} + (\mX - \mP)\va, 
 \]
where the $(i,j)$th entry of $\mP$, i.e., $P_{ij} = \pi_{ij}.$
 
Next, to compute the derivative of $\mathcal{L}_1(\vc,\mF)$ with respect to $\mF$, we have ${\partial \|\mF\|_F^2}/{\partial \mF} = 2\mF$ and
\begin{align*}
 	\frac{\partial\mathcal{T}_1}{\partial F_{uv}} &= \sum_{i=1}^{N}\sum_{k=1}^{D}\left({\frac{x_{ik}}{\pi_{ik}} - \frac{1-x_{ik}}{1-\pi_{ik}}}\right) \frac{\partial \pi_{ik}}{\partial F_{uv}} \nonumber \\
 	&= \sum_{i=1}^{N}(x_{iu} - \pi_{iu})C_{iv} \nonumber
\end{align*}
as 
 \begin{align}
 \frac{\partial \pi_{ik}}{\partial F_{uv}} 
&= \begin{cases}
			C_{iv}\pi_{ik}(1-\pi_{ik}), & \text{if $u=k$,}\\
            0, & \text{otherwise,}
		 \end{cases} 
 \end{align}
where recall that $\mC=[\vc,\mathbf{1}]\in \mathbb{R}^{N\times2}$. Then
\begin{align*}
 	\frac{\partial\mathcal{L}_1}{\partial \mF} 
 	&= (\mX - \mP)^T\mC - 2\alpha\mF. \nonumber
\end{align*}

\subsection{\texttt{GA-Affine-Real}}
The objective function in $(\cP2)$ can be written as
\[
\mathcal{L}_2(\vc,\mF) = \sum\limits_{j=1}^{N} 2\vc\rT|\bTheta_{:j}| - \underbrace{\|\mX - \vc\va\rT - {\mathbf 1}\vb\rT\|_F^2}_{= \|\mX - \mC\mF\rT\|_F^2} - \alpha\|\mF\|^2.
\]
Then derivative of $\mathcal{L}_2(\vc,\mF)$ with respect to $\vc$ is 
\begin{align}
{\frac{\partial \mathcal{L}_2(\vc,\mF)}{\partial \vc}} &= 2\lvert\bTheta\rvert\mathbf{1} + 2(\mX - \mC\mF\rT)\va
\nonumber
\end{align}
and the derivative of $\mathcal{L}_2(\vc,\mF)$ with respect to $\mF$ is
\begin{align}
{\frac{\partial \mathcal{L}_2(\vc,\mF)}{\partial \mF}} &=  2(\mX - \mC\mF\rT)\rT\mC - 2\alpha\mF.
\nonumber
\end{align}


\subsection{\texttt{GA-Nonlinear}}
The objective function in $(\cP3)$ can be written as 
\[
\mathcal{L}_3(\vc) = \sum\limits_{j=1}^{N} 2\vc\rT|\bTheta_{:j}| + \log {\rm det} \mathbf{K}(\vc) - \tr(\mS\mathbf{K}(\vc)).
\]
While the derivative of the first term with respect to $\vc$ is given in \eqref{eq:gradient_priorc}, we have
%
\begin{align}
	\frac{\partial {\rm log}{\rm det}{\mathbf{K}}}{\partial \vc} = \left[\tr\left(\mathbf{K}^{-1}\frac{\partial \mathbf{K}}{\partial c_1}\right),\cdots,\tr\left(\mathbf{K}^{-1}\frac{\partial \mathbf{K}}{\partial c_N}\right)\right]\rT \nonumber
	\label{eq:partial logdetk/c 1}
\end{align}
and 
\begin{align}
	\frac{\partial \tr\left(\mS\mathbf{K}\right)}{\partial \vc} = \left[\tr\left(\mS\frac{\partial \mathbf{K}}{\partial c_1}\right),\cdots,\tr\left(\mS\frac{\partial \mathbf{K}}{\partial c_N}\right)\right]\rT, \nonumber
\end{align}
where ${\partial \mathbf{K}}/{\partial c_i}$ is an $N\times N$ matrix for each $i=1,\cdots,N$, and its $(u,v)$th entry is given by
\begin{align}
	\left[\frac{\partial \mathbf{K}}{\partial c_i}\right]_{uv} = 
	\begin{cases}
			-1, & \text{if $u=i$ and $u\neq v$ or  $v=i$ and $v\neq u$,}\\
			-2, & \text{if $u=v=i$}, \\
            0, & \text{otherwise.}
		 \end{cases} 
\nonumber
\end{align}
For a symmetric matrix $\mA \in \mathbb{R}^{N \times N}$, we have 
\begin{align}
	\left[\mA\frac{\partial \mathbf{K}}{\partial c_i}\right]_{jj} = 
	\begin{cases}
			-A_{ji}, & \text{if $j\neq i$,}\\
			-\sum\limits_{j=1}^{N} A_{ij} - A_{ii}, & \text{if $j=i$}
		 \end{cases} 
\nonumber
\end{align}
with
\begin{align}
    \tr\left(\mA\frac{\partial \mathbf{K}}{\partial c_i}\right) &= -\sum\limits_{j=1,j\neq i}^{N} A_{ji}- \sum\limits_{j=1}^{N}A_{ij} - A_{ii} \nonumber\\
    &= -\sum\limits_{j=1}^{N}A_{ij} - \sum\limits_{j=1}^{N}A_{ji} = -2 \mathbf{1}\rT\mA_{i:}, \nonumber
\end{align}
where $\mA_{i:} = \left[A_{i1},A_{i2},\cdots,A_{iN} \right]\rT$. 
Since $\mathbf{K}^{-1}$ and $\mS$ are symmetric matrices, the derivative of $\mathcal{L}_3(\vc)$ with respect to $\vc$ is given by 
$$
{\frac{\partial \mathcal{L}_3(\vc)}{\partial \vc}} = 2\lvert\bTheta\rvert\mathbf{1} - 2\mathbf{K}^{-1}\mathbf{1} + 2\mS\mathbf{1}.
$$


\bibliographystyle{IEEEtran}
\bibliography{references}

\begin{thebibliography}{10}
\providecommand{\url}[1]{#1}
\csname url@samestyle\endcsname
\providecommand{\newblock}{\relax}
\providecommand{\bibinfo}[2]{#2}
\providecommand{\BIBentrySTDinterwordspacing}{\spaceskip=0pt\relax}
\providecommand{\BIBentryALTinterwordstretchfactor}{4}
\providecommand{\BIBentryALTinterwordspacing}{\spaceskip=\fontdimen2\font plus
\BIBentryALTinterwordstretchfactor\fontdimen3\font minus
  \fontdimen4\font\relax}
\providecommand{\BIBforeignlanguage}[2]{{%
\expandafter\ifx\csname l@#1\endcsname\relax
\typeout{** WARNING: IEEEtran.bst: No hyphenation pattern has been}%
\typeout{** loaded for the language `#1'. Using the pattern for}%
\typeout{** the default language instead.}%
\else
\language=\csname l@#1\endcsname
\fi
#2}}
\providecommand{\BIBdecl}{\relax}
\BIBdecl

\bibitem{barbera2015the}
P.~Barbera, N.~Wang, R.~Bonneau, J.~T. Jost, J.~Nagler, J.~Tucker, and
  S.~González-Bailón, ``The critical periphery in the growth of social
  protests.'' \emph{PloS one}, vol.~10, no.~11, p. e0143611, Nov. 2015.

\bibitem{Verma2016Emergence}
T.~Verma, F.~Russmann, N.~Araújo, J.~Nagler, and H.~Herrmann, ``Emergence of
  core–peripheries in networks.'' \emph{Nat. Commun.}, vol.~7, no.~1, pp.
  1--7, Jan. 2016.

\bibitem{bassett2013task}
D.~Bassett, N.~Wymbs, M.~Rombach, M.~Porter, P.~Mucha, and S.~Grafton,
  ``Task-based core-periphery organization of human brain dynamics.''
  \emph{PLoS Comput. Biol.}, vol.~9, no.~9, p. e1003171, Sep. 2013.

\bibitem{Harlalka2019Atypical}
V.~Harlalka, R.~Bapi, P.~Vinod, and D.~Roy, ``Atypical flexibility in dynamic
  functional connectivity quantifies the severity in autism spectrum
  disorder.'' \emph{Front. Hum. Neurosci.}, vol.~13, no.~6, Feb. 2019.

\bibitem{Kitsak2010Identification}
M.~Kitsak, L.~Gallos, S.~Havlin, F.~Liljeros, L.~Muchnik, H.~Stanley, and
  H.~Makse, ``Identification of influential spreaders in complex networks.''
  \emph{Nat. Phys.}, vol.~6, no.~11, pp. 888--93, Nov. 2010.

\bibitem{leskovec2012Learning}
\BIBentryALTinterwordspacing
J.~Leskovec and J.~Mcauley, ``Learning to discover social circles in ego
  networks.'' \emph{Adv. Neural. Inf. Process. Syst.}, 2012 (accessed Mar.,
  2022). [Online]. Available: \url{https://snap.stanford.edu/data/}
\BIBentrySTDinterwordspacing

\bibitem{Borgatti2000Models}
S.~Borgatti and M.~Everett, ``Models of core/periphery structures.'' \emph{Soc.
  Netw.}, vol.~21, no.~4, pp. 375--95, Oct. 2000.

\bibitem{Rombach2014Core}
M.~Rombach, M.~Porter, J.~Fowler, and P.~Mucha, ``Core-periphery structure in
  networks.'' \emph{SIAM J. Appl. Math.}, vol.~74, no.~1, pp. 167--90, Oct.
  2014.

\bibitem{Boyd2010Computing}
J.~Boyd, W.~Fitzgerald, M.~Mahutga, and D.~Smith, ``Computing continuous
  core/periphery structures for social relations data with {MINRES/SVD}.''
  \emph{Soc. Netw.}, vol.~32, no.~2, pp. 125--37, May 2010.

\bibitem{Alvarez2005K}
J.~Alvarez-Hamelin, L.~Dall'Asta, A.~Barrat, and A.~Vespignani, ``K-core
  decomposition of internet graphs: hierarchies, self-similarity and
  measurement biases.'' \emph{arXiv preprint cs/0511007}, Nov 2005.

\bibitem{Della2013Profiling}
R.~Della, F.~Dercole, and P.~C, ``Profiling core-periphery network structure by
  random walkers.'' \emph{Sci. Rep.}, vol.~3, no.~1, pp. 1--8, Mar. 2013.

\bibitem{Jia2019Random}
J.~Jia and A.~Benson, ``Random spatial network models for core-periphery
  structure.'' in \emph{Proc. ACM Int. Conf. on Web Search Data Mining}, New
  Orleans, USA, Jan. 2012.

\bibitem{Friedman2008Sparse}
J.~Friedman, T.~Hastie, and R.~Tibshirani, ``Sparse inverse covariance
  estimation with the graphical lasso.'' \emph{Biostat.}, vol.~9, no.~3, pp.
  432--41, Jul. 2008.

\bibitem{Gurugubelli2022Learning}
S.~Gurugubelli and S.~P. Chepuri, ``Learning sparse graphs with a
  core-periphery structure.'' in \emph{Proc. of the IEEE Int. Conf. on
  Acoustics, Speech and Signal Process. (ICASSP)}, Singapore, May 2022.

\bibitem{Hsieh2014QUIC}
C.~Hsieh, M.~Sustik, I.~Dhillon, and P.~Ravikumar, ``{QUIC}: quadratic
  approximation for sparse inverse covariance estimation.'' \emph{J. Mach.
  Learn. Res.}, vol.~15, no.~1, pp. 2911--47, Oct. 2014.

\bibitem{Karmarkar1984A}
N.~Karmarkar, ``A new polynomial-time algorithm for linear programming.'' in
  \emph{Proc. Annu. ACM Symp. Theory Comput.}, Washington, D.C., Dec. 1984.

\bibitem{Sen2008Collective}
\BIBentryALTinterwordspacing
P.~Sen, G.~Namata, M.~Bilgic, L.~Getoor, B.~Galligher, and T.~Eliassi-Rad,
  ``Collective classification in network data.'' \emph{AI Mag.}, 2008 (accessed
  Aug., 2021). [Online]. Available:
  \url{https://snap.stanford.edu/data/C-elegans-frontal.html}
\BIBentrySTDinterwordspacing

\bibitem{Greene2013Producing}
D.~Greene and P.~Cunningham, ``Producing a unified graph representation from
  multiple social network views.'' in \emph{Proc. ACM Int. Conf. on Web Sci.
  Conf.}, Paris, Frace, May. 2013.

\bibitem{Freeman1979the}
\BIBentryALTinterwordspacing
S.~Freeman and L.~Freeman, ``The networkers network: A study of the impact of a
  new communications medium on sociometric structure.'' \emph{School of Social
  Sciences University of Calif.}, 1979 (accessed Mar., 2022). [Online].
  Available: \url{https://toreopsahl.com/datasets/\#FreemansEIES}
\BIBentrySTDinterwordspacing

\bibitem{Cross2004hidden}
\BIBentryALTinterwordspacing
R.~Cross and A.~Parker, ``The hidden power of social networks.'' \emph{Harvard
  Business School Press, Boston, MA.}, 2004 (accessed Mar., 2022). [Online].
  Available: \url{https://toreopsahl.com/datasets/\#Cross\_Parker}
\BIBentrySTDinterwordspacing

\bibitem{Wu2020A}
Z.~Wu, S.~Pan, F.~Chen, G.~Long, C.~Zhang, and S.~Philip, ``A comprehensive
  survey on graph neural networks.'' \emph{IEEE Trans. Neural Netw. Learn.
  Syst.}, vol.~32, no.~1, pp. 4--24, Mar. 2020.

\bibitem{Hamilton20017Inductive}
W.~Hamilton, Z.~Ying, and J.~Leskovec, ``Inductive representation learning on
  large graphs.'' in \emph{Adv. Neural. Inf. Process. Syst.}, CA, USA, Dec.
  2017.

\bibitem{Dickstein2014neural}
S.~Dickstein, K.~Bannon, C.~Xavier, and M.~Milham, ``The neural correlates of
  attention deficit hyperactivity disorder: An ale meta‐analysis.'' \emph{J.
  Child Psychol. Psychiatry}, vol.~47, no.~10, pp. 1051--62, Nov. 2006.

\end{thebibliography}
\end{document}